\documentclass{article}

\PassOptionsToPackage{numbers,sort}{natbib}
\usepackage[preprint]{main}

\usepackage[utf8]{inputenc} 
\usepackage[T1]{fontenc}    
\usepackage{hyperref}       
\usepackage{url}            
\usepackage{booktabs}       
\usepackage{amsfonts}       
\usepackage{nicefrac}       
\usepackage{microtype}      
\usepackage{xcolor}         
\usepackage{array}
\usepackage{graphicx} 
\usepackage{subcaption}
\usepackage[percent]{overpic}
\usepackage{enumitem}
\usepackage{capt-of}
\usepackage{algorithm}
\usepackage{algpseudocode}
\usepackage{comment}
\usepackage{titletoc}
\usepackage{amssymb}
\usepackage[most]{tcolorbox}
\setcitestyle{numbers,square}
\algrenewcommand\algorithmicrequire{\textbf{Input:}}
\algrenewcommand\algorithmicensure{\textbf{Output:}}

\title{Reflection with Action-Induced Visual Differences for Desktop GUI Agents}

\author{ Yijie Ma, Chaoyue Niu, Fan Wu, Guihai Chen \\
Shanghai Jiao Tong University\\
}

\begin{document}

\maketitle

\begin{abstract}
    
    The Planner-Operator-Reflector (POR) framework is widely used in GUI agents to maintain objective alignment in complex tasks through modular collaboration. However, desktop GUIs introduce a key challenge: large, dense interfaces often exhibit subtle or scattered state changes, placing most of the burden on the reflector, which must compare pre- and post-action screens, while the planner and operator reason over a single state. Existing reflectors collapse change detection and outcome verification into one step, leaving evidence implicit and yielding weakly grounded decisions. To address this limitation, we propose Evidence-First Reflection (EFR), a two-stage reflector that explicitly decouples action-induced visual differences extraction from outcome verification. EFR identifies the action location and candidate changed regions with Set-of-Marks annotations, describes and filters action-relevant changes, and makes the final judgment from the cleaned evidence. This evidence-reasoning decoupled design makes reflection better grounded in screen transitions, while reducing both visual search complexity and reasoning burden. Experiments on OSWorld-Verified and WindowsAgentArena demonstrate that EFR improves reflector accuracy by 7.11\%, yielding average end-to-end task success gains of 5.94\% and 4.95\% on the two benchmarks, respectively.
\end{abstract}

\section{Introduction}
    
    Graphical User Interface (GUI) agents \citep{wang2024guiagentssurvey,zhang2025appagent,wang2024mobileagent,gao2023assistgui,rawles2025androidworld,niu2024screenagent,zhang2025ufo,agashe2024agents,xu2025aguvis} are designed to execute user instructions by interacting with digital interfaces in a human-like manner, typically leveraging vision-language models (VLMs). By unifying the observation space into screenshots and the action space into human-like operations, GUI agents can handle diverse tasks across both mobile and desktop environments \citep{xu2025aguvis,cheng2024seeclick}. This broader task scope introduces higher complexity, motivating existing studies to adopt the mainstream Planner-Operator-Reflector (POR) framework \citep{ye2025mobileagentv3,wang2025mobileagente,agashe2024agents,agashe2025agents2,li2025mobileuse}, which decomposes agent control into collaborative modules to maintain task alignment and reduce the burden on any single module. The planner formulates a high-level plan, the operator executes the corresponding low-level action, and the reflector assesses whether the action has successfully achieved the intended goal. Serving as a critical safeguard in this workflow, the reflector helps ensure action accuracy and prevents error accumulation over steps. It typically performs this assessment in a single inference pass by observing the pre- and post-action screenshots and comparing the observation with the intended action to generate a judgment \citep{wang2024mobileagentv2,ye2025mobileagentv3,wang2025mobileagente,agashe2025agents2,liu2025pcagent}.

\begin{figure}[t]
        \centering
        \begin{subfigure}[t]{0.485\textwidth}
            \centering
            \includegraphics[width=\linewidth,keepaspectratio]{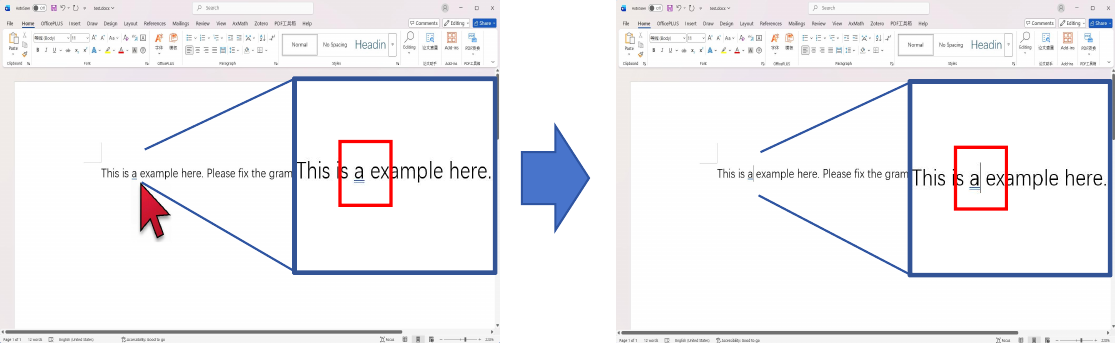}
            \caption{Only a cursor change.}
            \label{fig:intro-problems-small}
        \end{subfigure}\hfill
        \begin{subfigure}[t]{0.485\textwidth}
            \centering
            \includegraphics[width=\linewidth,keepaspectratio]{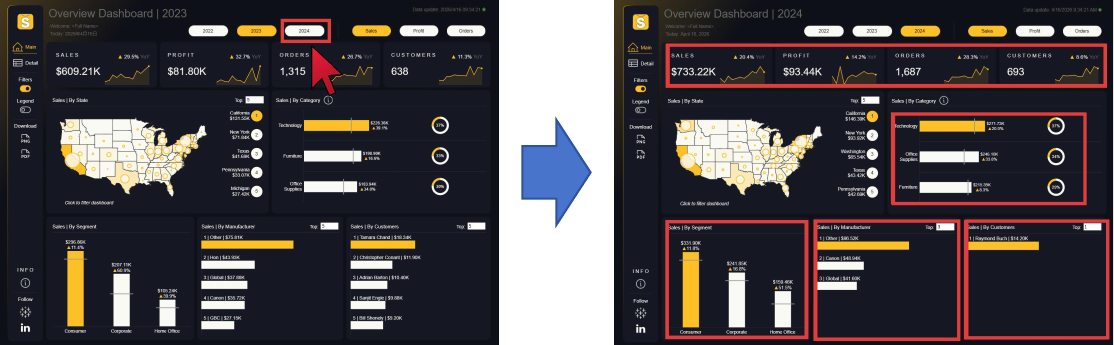}
            \caption{Many page regions change simultaneously.}
            \label{fig:intro-problems-big}
        \end{subfigure}
    
        \caption{Two representative examples of interface changes in desktop environments. Fig.~\ref{fig:intro-problems-small}: the action outcome is indicated by a subtle cursor movement after clicking the position of the letter ``a'', making the change easy to overlook. Fig.~\ref{fig:intro-problems-big}: switching the year triggers extensive updates across the interface, making it challenging to organize all relevant evidence to assess the action's effect.}\label{fig:intro-problems}
    \end{figure}

    However, reflection in desktop environments is generally more challenging than in mobile environments, as desktop screens are often several to even dozens of times larger. Such expansive interfaces tend to exhibit more complex layouts and a richer set of elements, making it difficult to comprehensively detect and interpret GUI changes \citep{zhang2025ufo,agashe2024agents,liu2025pcagent,qin2025uitars}. This challenge disproportionately affects the reflector compared to the planner and operator: while the latter mainly need to reason about the current interface state, the reflector must jointly analyze both pre- and post-action screenshots to determine whether an action has succeeded. Figure~\ref{fig:intro-problems} shows two representative examples: one where the action outcome is extremely subtle, and the other where the action induces multiple, spatially distributed changes. In both cases, the reflector struggles to capture and organize the complete set of interface changes, leading it to judge the action outcome based on only a limited and potentially incorrect subset of changes. In our experiment with GUI-Owl-32B trajectories using the conventional POR framework on OSWorld-Verified \citep{xie2024osworld,osworld_verified}, 61 out of 223 failed trajectories (27.35\%) are caused by incorrect reflector judgments. Among these reflection failures, 43 cases (70.49\%) stem from page difference identification errors, while 18 cases (29.51\%) result from success judgment errors given correctly identified page differences, as detailed in Appendix Table~\ref{tab:appendix-reflector-failure-causes}. This indicates that the conventional reflector is not sufficiently reliable in complex desktop environments.

    \begin{figure}[t]
        \centering
        \begin{subfigure}[t]{0.34\textwidth}
            \centering
            \includegraphics[height=0.13\textheight,keepaspectratio]{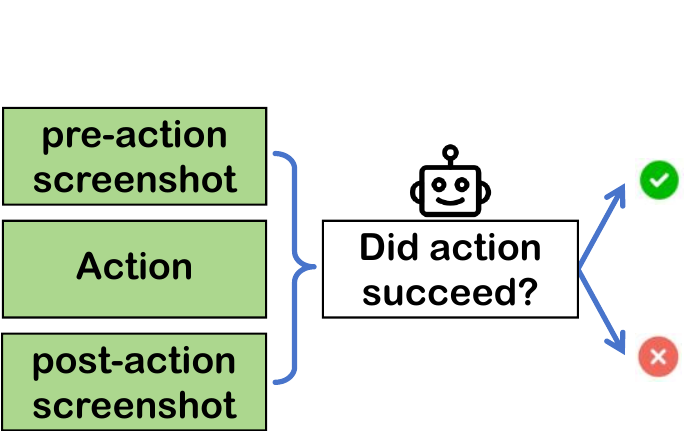}
            \caption{Conventional Reflector.}
            \label{fig:reflector-comparison-conventional}
        \end{subfigure}\hfill
        \begin{subfigure}[t]{0.65\textwidth}
            \centering
            \includegraphics[height=0.13\textheight,keepaspectratio]{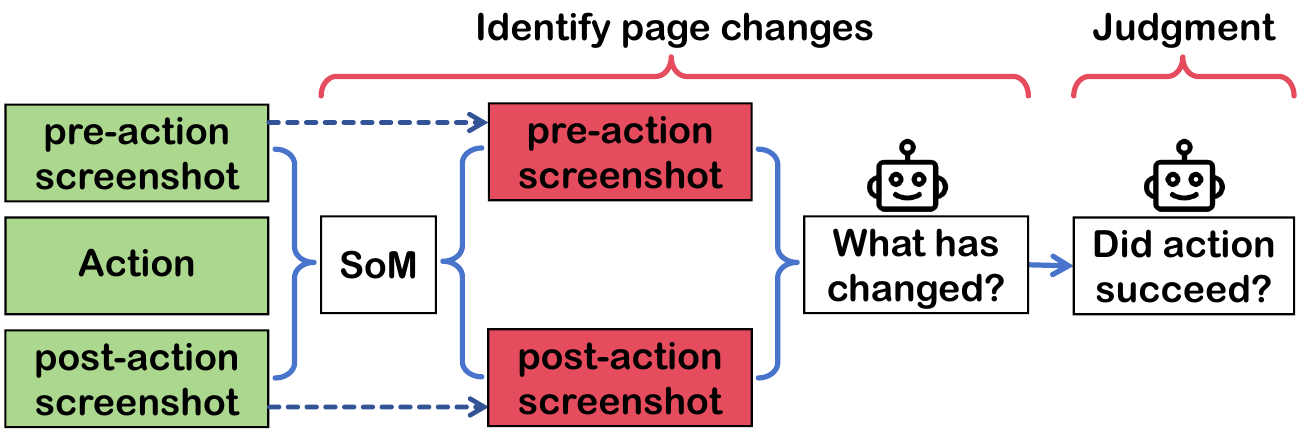}
            \caption{EFR.}
            \label{fig:reflector-comparison-efr}
        \end{subfigure}
        \caption{The conventional reflector performs one-step reasoning to judge action success directly, whereas EFR first describes page changes from SoM annotations as explicit evidence and then verifies the action based on the extracted evidence.}
        \label{fig:reflector-comparison}
    \end{figure}
    
    By deconstructing the logic of conventional reflectors, we find that they always couple two operations with different roles: first establishing what changed as observable evidence, and then interpreting whether the evidence supports a successful action. When these operations are coupled into one inference step, the evidence used for the final judgment remains implicit and cannot be checked before the VLM reaches a conclusion. To address this limitation, we revisit the internal design of the reflector itself and propose \textbf{Evidence-First Reflection (EFR)}, a two-stage reflector that explicitly decouples visual evidence extraction from action outcome verification. It mitigates potential bias by decoupling the objective observation of page changes from subjective assumptions regarding intended action outcomes, while reducing the visual burden by repurposing Set-of-Marks (SoM), originally used to label candidate page elements, to annotate page changes before and after the action. Figure~\ref{fig:reflector-comparison} shows the core difference between conventional reflector and EFR. Specifically, EFR first leverages SoM to annotate action-induced change regions and the grounded action location, prompting the VLM to objectively describe and filter action-relevant page changes as evidence. Then, it judges whether the action succeeded based on the cleaned evidence and the intended action. By separating ``what changed'' from ``whether the action succeeded'', EFR reduces both the visual burden and the reasoning burden of reflection. In summary, the key contributions include:
    \begin{itemize}[leftmargin=*, labelsep=0.5em]
        \item We observe that the inherent complexity of desktop environments substantially complicates change identification, thereby compromising the reliability of conventional reflectors that infer action success directly in a single reasoning step.
        \item We reconstruct the reflector as a two-stage sequential process that explicitly separates page difference identification from action success judgment. The first stage describes and organizes action-relevant visual changes from SoM annotations, and the second stage verifies action success based on the extracted evidence, making the dependency between evidence construction and outcome judgment explicit. This design grounds the final judgment in a checkable evidence chain and reduces interference between visual search and outcome reasoning.
        \item We implement the POR framework following Mobile-Agent-v3 \citep{ye2025mobileagentv3} and evaluate EFR with both GUI-specialized and general backbones on OSWorld-Verified \citep{xie2024osworld,osworld_verified} and WindowsAgentArena \citep{bonatti2025windowsagentarena} benchmarks. Averaged over three runs, EFR improves end-to-end task success rates across all model-benchmark settings, with average gains of 5.94\% and 4.95\%, respectively. Further analysis demonstrates that the performance gains are strongly correlated with improvements in reflector accuracy. Ablation studies validate the necessity of each module. Notably, the performance improvement comes with moderate per-task latency overheads, averaging 15.8\% and 13.3\%, respectively.
    \end{itemize}

\section{Related Work}
\label{sec:related-work}

    \textbf{POR Framework.} The key of POR framework is to explicitly decouple planning, execution, and feedback verification \citep{shinn2023reflexion,madaan2023selfrefine,wang2024mobileagentv2}. This paradigm shares a lineage with the reasoning-acting loop exemplified by ReAct \citep{yao2022react}. Many studies adopt this paradigm and further refined it to better handle complex tasks. For instance, Mobile-Agent-v3 \citep{ye2025mobileagentv3} introduces Planner-Operator-Reflector-Notetaker framework to handle multi-step navigation and error correction. Mobile-Agent-E \citep{wang2025mobileagente} and MobileUse \citep{li2025mobileuse} enhance hierarchical planning and introduce an on-demand, reflection-driven workflow. Agent-S2 \citep{agashe2025agents2} uses dynamic collaboration to plan various modules. While these studies largely retain the POR framework as-is and enhance it with additional components or strategies, our work revisits its internal modular design by reconstructing the reflector module.

    \textbf{Reflector.} Reflector is typically used to convert failures into reusable memory or to support hierarchical closed-loop correction \citep{shinn2023reflexion,madaan2023selfrefine,du2024anytool}. In GUI agents, the use of reflectors has been further extended within the overall agent pipeline: MobileUse \citep{li2025mobileuse} introduces hierarchical reflection to identify and correct errors across different granularities, InfiGUIAgent \citep{liu2025infiguiagent} and D-Artemis \citep{mi2025dartemis} perform reflection at different stages of the action execution process, and BacktrackAgent \citep{wu2025backtrackagent} leverages reflection for trajectory rollback. Their improvements are largely about deploying the reflector in different roles or stages of the agent pipeline. As a result, most approaches still implicitly assume that the reflector can operate as a single-step module. In contrast, our work focuses on its internal reasoning logic and improves it by explicitly decoupling evidence extraction from success judgment.

    \textbf{SoM in GUI Agent.} SoM annotates interface content by overlaying explicit markers, such as letters, numbers, or bounding boxes, on screenshots \citep{yang2023som}. In prior GUI agents, it was used to identify candidate elements before taking actions \citep{cheng2024seeclick,li2024mug,lu2024omniparser,gou2024uground,yang2025ariaui,li2025autogui}. For instance, AppAgent \citep{zhang2025appagent} overlays semi-transparent numeric markers onto candidate elements parsed from XML on mobile screenshots, enabling the model to execute operations using numeric indices rather than raw coordinates. UFO \citep{zhang2025ufo} simultaneously provides the agent with a raw screenshot and an annotated screenshot, encoding UI controls with numbers and colors. In addition to these specific methods, environments like VisualWebArena \citep{koh2024visualwebarena} and OSWorld \citep{xie2024osworld} also employ standardized element-level SoM observation representations to facilitate element recognition. In contrast, we extend SoM beyond target-element localization and apply it to post-action page-change identification within the reflection module.

\section{Problem Formulation}
\label{Problem Formulation}
    We first describe the workflow of the conventional POR framework, which employs a closed-loop iterative process to complete tasks. Within a single execution round, the Planner first formulates a comprehensive plan and decomposes it into discrete sub-goals based on the current state and overall task goal. Subsequently, the Operator receives the active sub-goal and generates a grounded action for execution. Finally, the Reflector verifies whether the executed action achieves the intended goal by comparing pre- and post-action screenshots. Formally, at each step $t$, the operations of the modules are defined as follows:

    \textbf{Planner.} Given the task goal $G$, the Planner leverages the current screenshot $S_t$ together with the previous state, comprising the action history $H_{<t}$ and the reflection from the preceding step $R_{t-1}$, to derive the next sub-goal $P_t$: 
    \begin{equation}
    P_t = f_{\mathrm{planner}}(G, S_t, H_{<t}, R_{t-1}).
    \end{equation}
    \textbf{Operator.} The Operator then grounds the sub-goal $P_t$ in the current visual observation $S_t$ and outputs the executable action $A_t$:
    \begin{equation}
    A_t = f_{\mathrm{operator}}(P_t, S_t).
    \end{equation}
    \textbf{Reflector.} Finally, after execution, the Reflector compares the pre-action screenshot $S_t^{pre}$ with the post-action screenshot $S_t^{post}$ to assess the outcome of the action $A_t$ and produce the reflection $R_t$:
    \begin{equation}
    R_t = f_{\mathrm{reflector}}(P_t, A_t, S_t^{pre}, S_t^{post}).
    \end{equation}
    Despite the modular decomposition of the POR framework, the reflection phase remains a monolithic challenge. To produce a reliable assessment, the Reflector must simultaneously address two distinct objectives: (1) identifying visual transformations between pre- and post-action screenshots, (2) determining whether these changes satisfy the intended goal. Conventional methodologies typically process these steps in a single pass by directly feeding the action description and both screenshots into the VLM. However, this design is brittle in desktop environments where screenshots are high-resolution and information-dense. Specifically, subtle changes are often overlooked, while intricate changes can lead to fragmented analysis and hasty conclusions. To address these limitations, it is essential to equip the reflector with the ability to comprehensively perceive the page changes, and to assess action correctness strictly on the basis of the observed state transition.
    
\section{Method}
\label{Method}
    \begin{figure}[t]
        \centering
        \includegraphics[width=\textwidth]{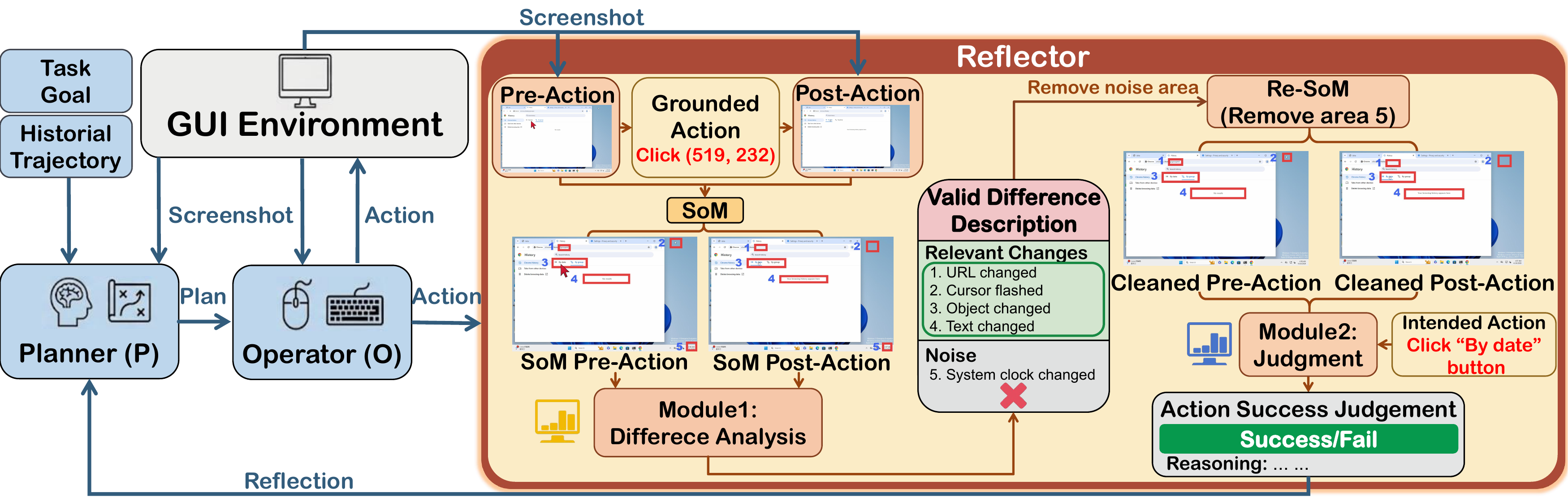}
        \caption{Overview of EFR in POR framework. Given the pre-action screenshot $S_{pre}$, post-action screenshot $S_{post}$, grounded action description $A_{ground}$, and intended action description $A_{intend}$, EFR first annotates candidate visual differences and the action location with SoM, then filters action-relevant evidence, and finally judges action success based on the cleaned evidence.}
        \label{fig:EFR}
    \end{figure}

    In this section, we introduce \textbf{EFR}, a two-stage evidence-based reflection framework designed to extract salient features under complex visual transformations and formulate judgments grounded in explicit evidence. Figure~\ref{fig:EFR} illustrates the overall framework, with detailed descriptions of each component provided in the subsequent subsections. Within the POR loop, EFR replaces the conventional reflector: after the planner generates a plan and the operator executes the corresponding action, EFR evaluates whether the action produced correct result and feeds the reflection result back into the next planning step. Rather than directly inferring action succeeded from raw pre- and post-action screenshots, EFR explicitly decouples visual change recognition from outcome verification. This approach is built upon a twofold strategy: first, we preprocess the screenshots to reduce the perceptual burden associated with detecting subtle or distributed interface changes; second, we implement a two-stage reasoning pipeline to mitigate the cognitive complexity of success evaluation. Specifically, EFR utilizes SoM to annotate both the grounded action location and visual differences between the screenshots, thereby constructing a foundation of explicit visual evidence. Then it systematically describes key changes and filters out environmental noise. By synthesizing this refined evidence with the intended goal, EFR determines the correctness of the executed action. Appendix~\ref{sec:step-level-reflection-example} presents an example, illustrating the comparison between the conventional reflector and EFR.

\subsection{Page Changes Annotated With SoM}

    Identifying page changes is a specialized perceptual task for which general-purpose VLMs often underperform, whereas task-specific computer vision methods are more effective and efficient. Instead of relying on the VLM to detect changes unaided, we first apply the SoM technique to explicitly annotate visual differences. Existing SoM methods were typically designed to assist target-element localization during action execution. We repurpose this idea for the reflection stage to support page-change analysis. This step explicitly converts dispersed differences between two screenshots into consolidated, structured evidence that the VLM can directly reference: changed regions are marked with red bounding boxes and indexed for region-level comparison. In addition, we annotate the action coordinates with blue circles to provide spatial context regarding where the action was executed. These explicit visual markers allow the VLM to directly locate salient changes on the page and reason over them as structured visual evidence.
    
    Given the pre-action screenshot $S_{pre}$, the post-action screenshot $S_{post}$, and the grounded action $A_{ground}$ which contains action type and coordinates, we first identify candidate changes through pixel-wise comparison. Nearby changed regions are then merged into coherent visual-change areas, which are finally mapped to rectangular bounding boxes. These boxes and their corresponding indices are then overlaid onto both screenshots. In addition, the action coordinates are annotated on the pre-action screenshot. This procedure yields two intermediate representations: $S_{pre}^{som}$, which depicts the pre-action screenshot marked with change areas and action coordinates, and $S_{post}^{som}$, which only displays change areas on the post-action screenshot:
    \begin{equation}
        (S_{pre}^{som}, S_{post}^{som}) = f_{\mathrm{som}}(S_{pre}, S_{post}, A_{ground}).
        \label{eq:som}
    \end{equation}
    By incorporating SoM, we explicitly indicate the locations of visual changes. An illustrative example is presented in Figure~\ref{fig:som-example}. Further details regarding the identification and annotation process are provided in the Appendix~\ref{sec:som-detail}. This approach reduces the perceptual load, thereby enabling the VLM to perform localized comparisons within the annotated regions.

    \begin{figure*}[t]
        \centering
        \begin{subfigure}[b]{0.4\textwidth}
            \centering
            \includegraphics[width=\linewidth]{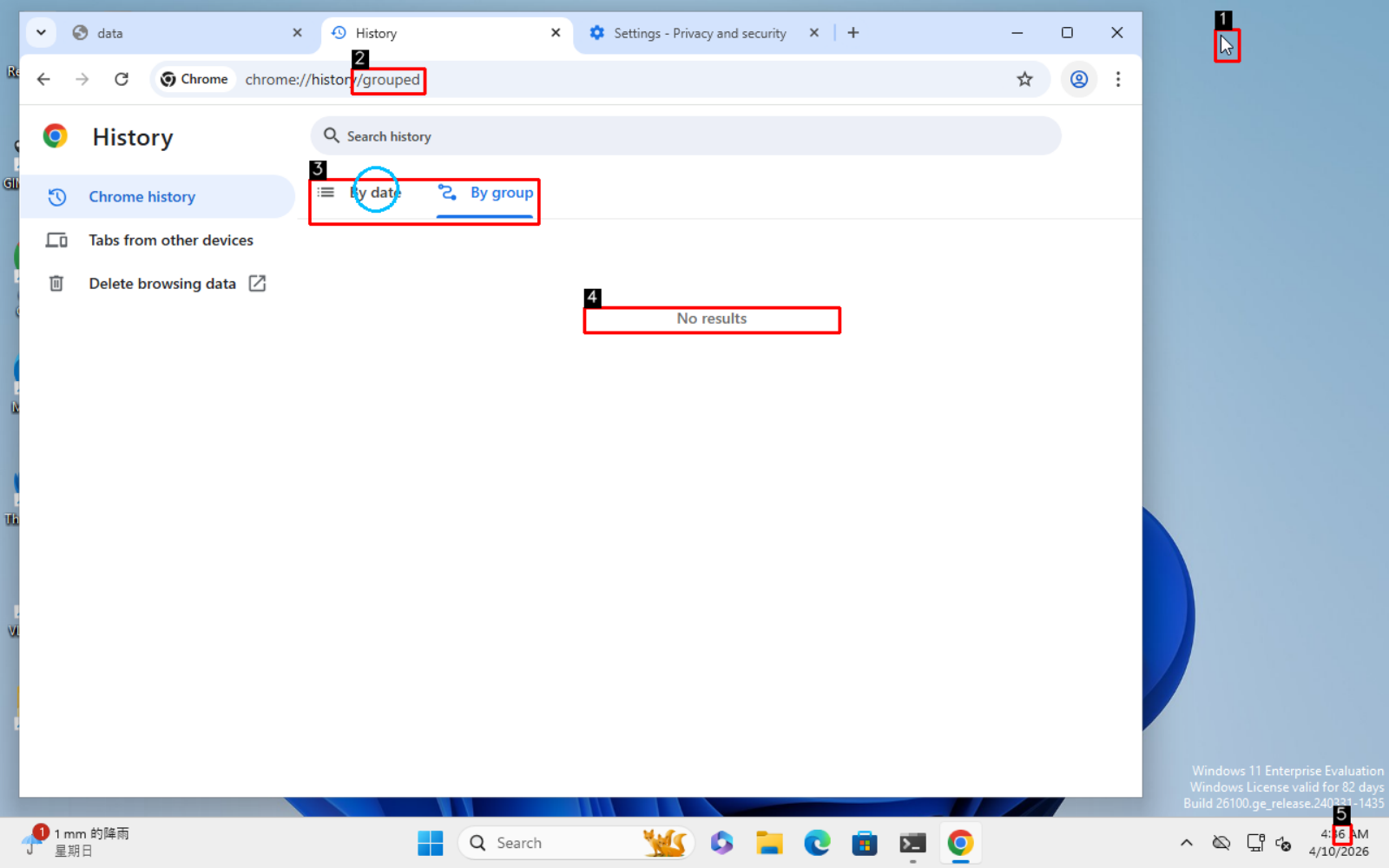}
            \caption{Pre-action marked screenshot}
            \label{fig:som-pre-trace}
        \end{subfigure}
        \hfill
        \begin{subfigure}[b]{0.4\textwidth}
            \centering
            \includegraphics[width=\linewidth]{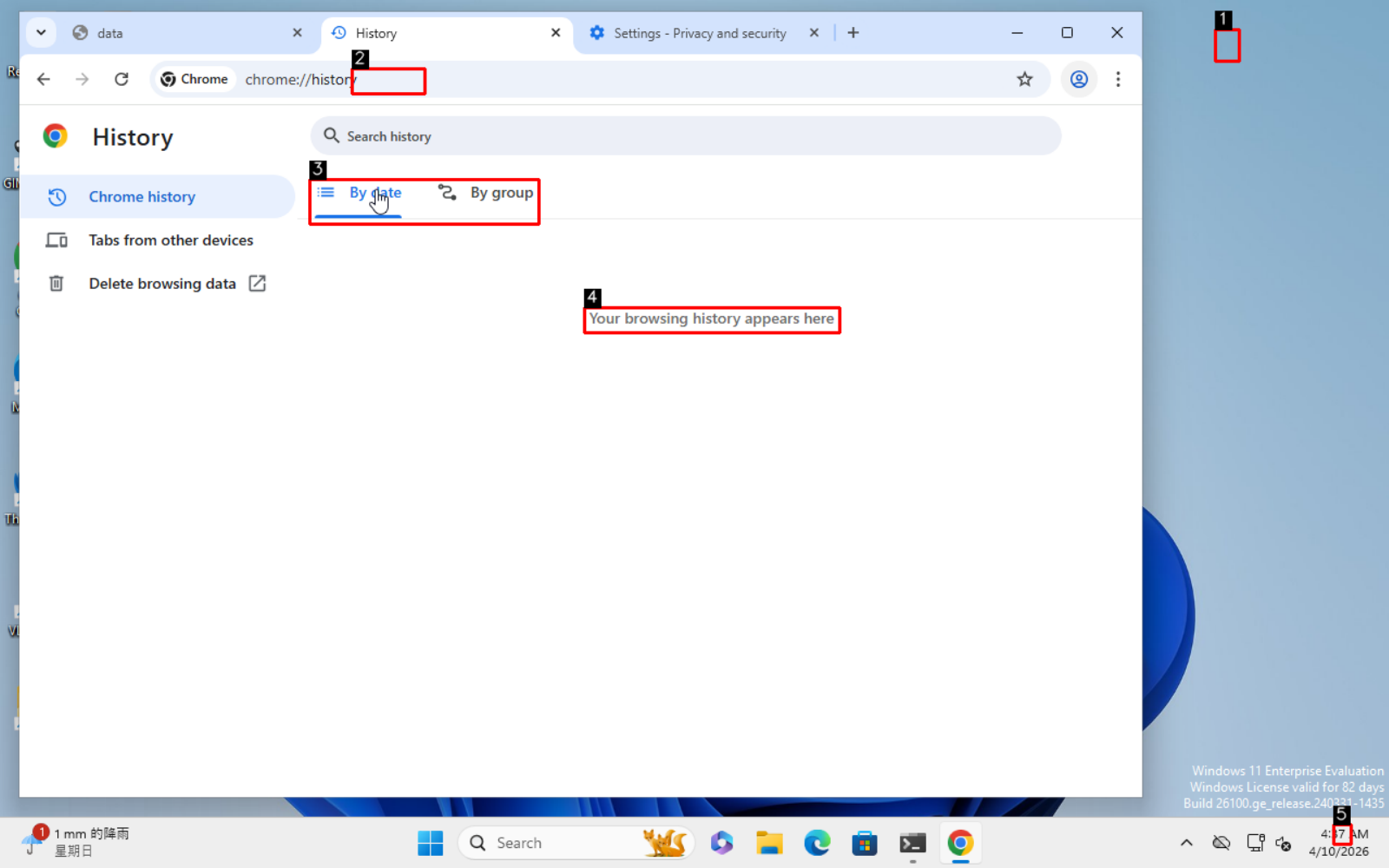}
            \caption{Post-action marked screenshot}
            \label{fig:som-post-trace}
        \end{subfigure}
        
        \caption{A concrete example of SoM annotation. Fig.~\ref{fig:som-pre-trace} shows the pre-action marked screenshot, where the executed action location and difference regions are annotated. Fig.~\ref{fig:som-post-trace} shows the post-action marked screenshot, where difference regions are annotated.}
        \label{fig:som-example}
    \end{figure*}

\subsection{Action-Induced Page Difference Analysis and Filtering}

\begin{figure}[t]
        \centering
        \setlength{\tabcolsep}{5pt}
        \renewcommand{\arraystretch}{1.0}
        \begin{tabular}{@{}c c c c c@{}}
            \begin{minipage}[c]{0.26\textwidth}
                \centering
                {\small \textbf{Marked Screenshots} $(S_{pre}^{som}, S_{post}^{som})$\par}
                \vspace{0.3em}
                {\scriptsize \textbf{Pre-action}\par}
                \includegraphics[width=\linewidth]{figures/som_example/pre_trace.pdf}\par
                \vspace{0.35em}
                {\scriptsize \textbf{Post-action}\par}
                \includegraphics[width=\linewidth]{figures/som_example/post_trace.pdf}
            \end{minipage}
            &
            {\Large $\Rightarrow$}
            &
            \begin{minipage}[c]{0.30\textwidth}
                \centering
                \fbox{%
                    \begin{minipage}[t]{0.93\linewidth}
                        \small
                        \textbf{Region-Level Descriptions}\par
                        \vspace{0.35em}
                        \textbf{Area 1} \textcolor{blue}{[Relevant]}: Mouse cursor disappears.\par
                        \textbf{Area 2} \textcolor{blue}{[Relevant]}: URL changes.\par
                        \textbf{Area 3} \textcolor{blue}{[Relevant]}: ``By group'' switches to ``By date''.\par
                        \textbf{Area 4} \textcolor{blue}{[Relevant]}: Content changes.\par
                        \textbf{Area 5} \textcolor{red}{[Noise]}: System clock changes.
                    \end{minipage}%
                }
            \end{minipage}
            &
            {\Large $\Rightarrow$}
            &
            \begin{minipage}[c]{0.26\textwidth}
                \centering
                {\small \textbf{Clean Evidence} $(S_{pre}^{clean}, S_{post}^{clean})$\par}
                \vspace{0.3em}
                {\scriptsize \textbf{Pre-action}\par}
                \includegraphics[width=\linewidth]{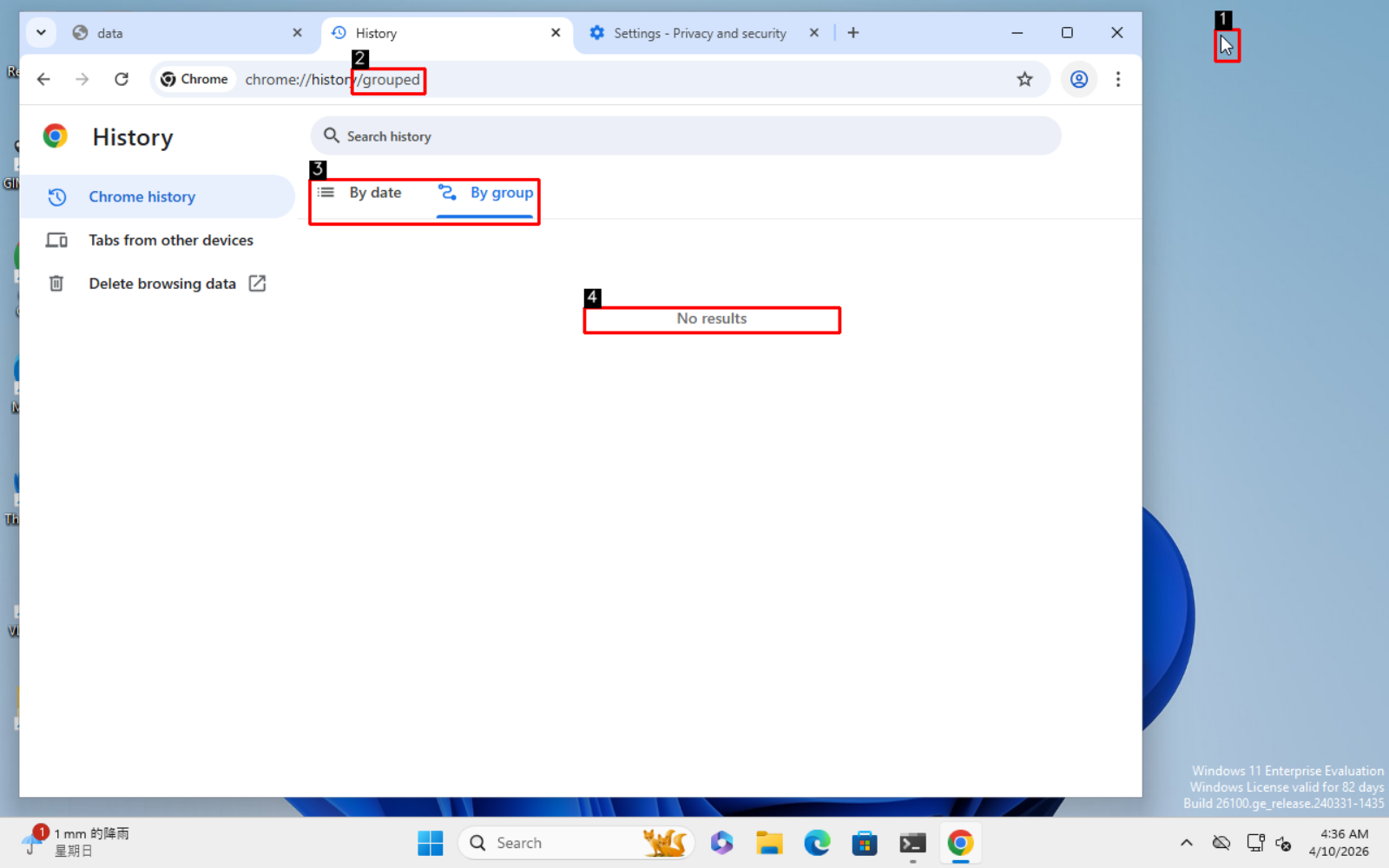}\par
                \vspace{0.35em}
                {\scriptsize \textbf{Post-action}\par}
                \includegraphics[width=\linewidth]{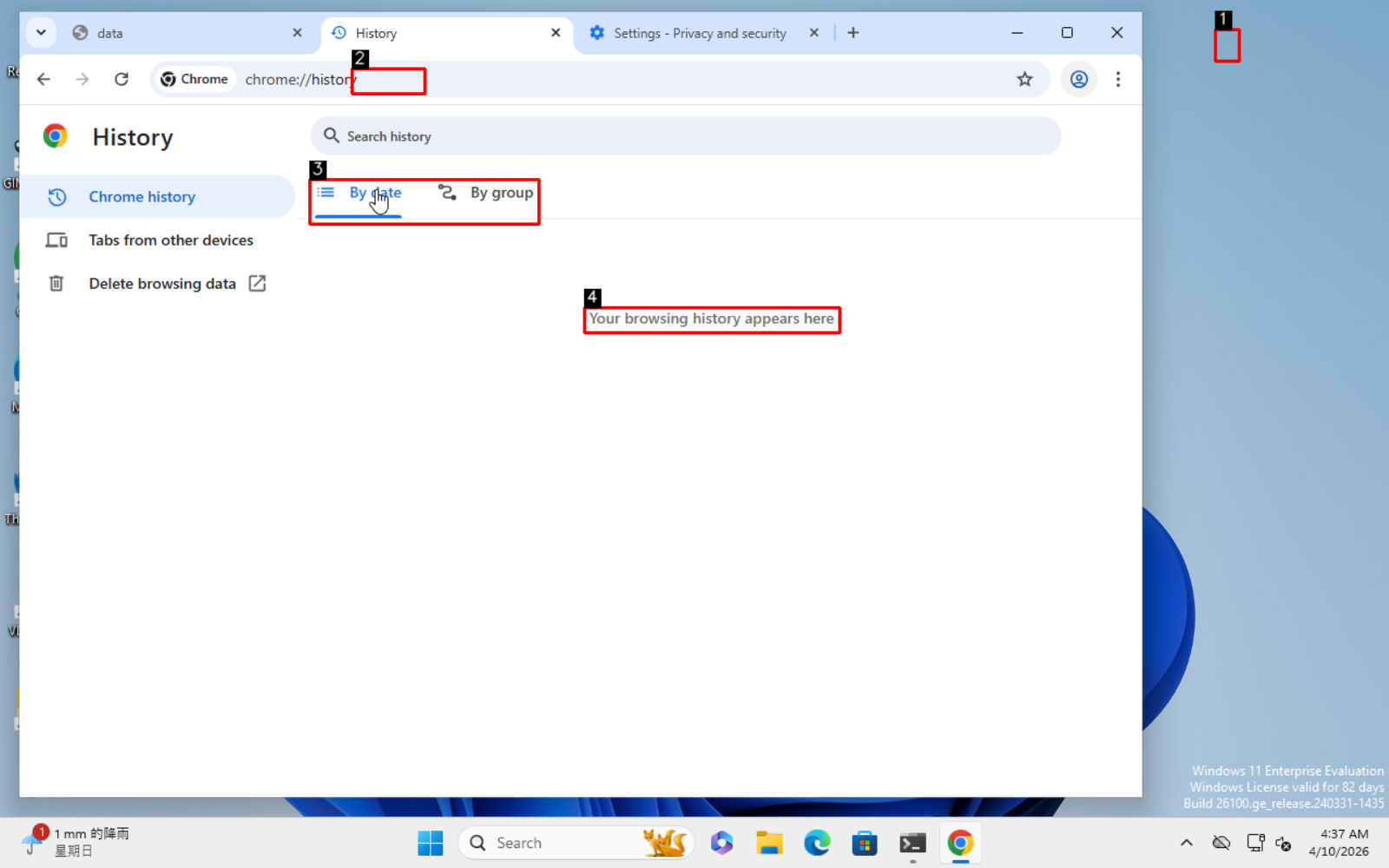}
            \end{minipage}
        \end{tabular}
        \caption{From total changes to clean evidence. The reflector first compares the annotated pre-action and post-action screenshots and outputs change descriptions together with relevant/noise tags. In this example, the bottom right Area 5 is filtered out as environment noise. The remaining relevant regions are then redrawn to form the clean evidence pair.}
        \label{fig:clean-evidence}
    \end{figure}

    After obtaining the annotated screenshots, the next phase involves an intermediate step aimed at extracting explicit visual evidence. An illustrative example is presented in Figure~\ref{fig:clean-evidence}. During this stage, the VLM is provided with the marked pre- and post-action screenshots ($S_{pre}^{som}, S_{post}^{som}$) and the grounded action $A_{ground}$. Importantly, $A_{ground}$ contains only the action type and its coordinates, rather than the original intended action $A_{intend}$. For example, the intended action may be ``click the button in the upper-left corner,'' whereas the grounded action is represented as ``click at coordinate $(x, y)$.'' This design encourages the reflector to extract evidence from the actual execution result rather than from the expected intention, reducing intention bias and improving the objectivity of the intermediate visual evidence.
    
    Given these inputs, the VLM generates explicit visual evidence in the form of region-level change descriptions. For each annotated region, it compares the corresponding visual contents in the marked pre- and post-action screenshots and independently describes the observed change. Since all detected changes are annotated, these regions may also contain incidental system noise, such as clock refreshes. Therefore, during this region-wise traversal, the VLM further distinguishes whether each described change reflects system noise or a meaningful state transition. Finally, it outputs the set of change descriptions $C$, categorized into system noise $\mathcal{N}$ and action relevant changes $\mathcal{E}$:
    \begin{equation}
        (C, \mathcal{N}, \mathcal{E}) = f_{\mathrm{desc}}(S_{pre}^{som}, S_{post}^{som}, A_{ground}).
        \label{eq:desc}
    \end{equation}
    We retain only the descriptions and markers associated with meaningful regional changes while filtering out system noise. This procedure requires re-labeling the original screenshots, $S_{pre}$ and $S_{post}$, to derive $S_{pre}^{clean}$ and $S_{post}^{clean}$, which isolate non-noise changes for use in the subsequent success judgment phase. We also filter out noise-related descriptions from $C$ and obtain $C^{clean}$:
    \begin{equation}
        (S_{pre}^{clean}, S_{post}^{clean}, C^{clean}) = f_{\mathrm{clean}}(S_{pre}, S_{post}, \mathcal{E}, C).
        \label{eq:clean}
    \end{equation}
    
\subsection{Action Outcome Verification Based on Evidence}

    Upon obtaining the filtered evidence, the final stage involves determining action success. At this step, the screenshots $S_{pre}^{clean}$ and $S_{post}^{clean}$, the filtered change description $C^{clean}$, and the original intended action $A_{intend}$ are jointly provided to the reflector. Based on this filtered visual evidence, the VLM evaluates whether the action has achieved the intended objective and outputs the reflection $R$, which consists of a success label, key evidence, and a status summary for downstream modules:
    \begin{equation}
    R = f_{\mathrm{judge}}(S_{pre}^{clean}, S_{post}^{clean}, C^{clean}, A_{intend}).
    \label{eq:judge}
    \end{equation}

\section{Evaluation}

\subsection{Evaluation Setup}

    \textbf{Benchmarks.} We use two benchmarks across diverse environments for computer-based agents: \textbf{OSWorld-Verified} \citep{xie2024osworld,osworld_verified} and \textbf{WindowsAgentArena} \citep{bonatti2025windowsagentarena}. OSWorld-Verified comprises 369 real-world Linux system tasks. In accordance with official guidelines and recent evaluations \citep{xie2024osworld,ye2025mobileagentv3,qin2025uitars,wang2025opencua}, we exclude 8 Google Drive-related tasks that require manual configuration, resulting in a 361-task subset. This evaluation spans 10 application domains, including Chrome, GIMP, LibreOffice, Thunderbird, VLC, and VSCode. WindowsAgentArena consists of 154 Windows system tasks encompassing various applications; we include the complete task set for our analysis. Appendix~\ref{sec:benchmark-grouping-details} shows the task type.

    \textbf{Baselines and Models.} We adopt the conventional POR framework as our baseline, adhering to the implementation established in Mobile-Agent-v3 \citep{ye2025mobileagentv3}. Our method replaces the standard reflector module with the proposed EFR variant. To ensure a comprehensive evaluation across diverse architectures, we select both GUI-specialized and general-purpose backbones from open-source and proprietary families. Specifically, we benchmark both POR and POR-EFR using four models: GUI-Owl-32B \citep{ye2025mobileagentv3}, Qwen3-vl-32B-instruct \citep{bai2025qwen3vl}, Kimi-K2.5 \citep{kimi2025k25blog}, and Seed-1.8 \citep{bytedanceseed2025seed18}.

    \textbf{Configurations.} All experiments are conducted within the Docker environments provided by the respective benchmarks. The agent operates on raw, screenshot-only observations without further preprocessing or structured UI metadata; all actions are executed via the \texttt{pyautogui} library, for details see Appendix~\ref{sec:action-space}. We set the maximum number of steps to 50 for GUI-specialized and proprietary models, while open-source general-purpose models are restricted to a maximum of 15 steps. We repeat each main POR and POR-EFR experiment three times and report the mean success rate across 3 runs in Tables~\ref{tab:osworld-results} and~\ref{tab:windowsagentarena-results}.

\subsection{Main Results and Analysis}
    \begin{table}[t]
        \centering
        \small
        \setlength{\tabcolsep}{4.5pt}
        \renewcommand{\arraystretch}{1.1}
        \caption{Mean success rates (\%) over three runs on the OSWorld-Verified benchmark. Rows correspond to task types. For each model backbone, the table reports the mean POR result, the mean EFR result, and their absolute difference $\Delta$ in percentage points. The best mean accuracy in each row is bolded; ties are bolded together.}
        \label{tab:osworld-results}
        \resizebox{\textwidth}{!}{%
        \begin{tabular}{@{}l*{12}{c}@{}}
            \toprule
            \textbf{Task Type} &
            \multicolumn{3}{c}{\textbf{GUI-Owl-32B \citep{ye2025mobileagentv3}}} &
            \multicolumn{3}{c}{\textbf{Qwen3-VL-32B \citep{bai2025qwen3vl}}} &
            \multicolumn{3}{c}{\textbf{Kimi-K2.5 \citep{kimi2025k25blog}}} &
            \multicolumn{3}{c}{\textbf{Seed-1.8 \citep{bytedanceseed2025seed18}}} \\
            \cmidrule(lr){2-4}\cmidrule(lr){5-7}\cmidrule(lr){8-10}\cmidrule(lr){11-13}
            & \textbf{POR} & \textbf{EFR} & \textbf{$\Delta$} & \textbf{POR} & \textbf{EFR} & \textbf{$\Delta$} & \textbf{POR} & \textbf{EFR} & \textbf{$\Delta$} & \textbf{POR} & \textbf{EFR} & \textbf{$\Delta$} \\
            \midrule
            OS & 61.11 & 62.50 & +1.39 & 56.94 & 55.56 & -1.38 & 59.72 & 68.06 & +8.34 & 59.72 & \textbf{70.83} & +11.11 \\
            Office & 30.16 & 33.62 & +3.46 & 23.02 & 26.72 & +3.70 & 50.97 & \textbf{61.20} & +10.23 & 47.23 & 51.25 & +4.02 \\
            Daily & 50.24 & 58.02 & +7.78 & 42.03 & 48.67 & +6.64 & 57.88 & \textbf{64.50} & +6.62 & 49.00 & 58.36 & +9.36 \\
            Professional & 64.63 & \textbf{80.95} & +16.32 & 66.67 & 73.47 & +6.80 & \textbf{80.95} & 78.91 & -2.04 & 62.59 & 63.27 & +0.68 \\
            Workflow & 16.92 & 24.60 & +7.68 & 15.48 & 21.51 & +6.03 & 32.50 & \textbf{41.66} & +9.16 & 25.32 & 25.78 & +0.46 \\
            \midrule
            Overall & 37.82 & 44.91 & +7.09 & 33.36 & 38.38 & +5.02 & 52.36 & \textbf{59.74} & +7.38 & 44.88 & 49.16 & +4.28 \\
            \bottomrule
        \end{tabular}%
        }
    \end{table}
    \begin{table}[t]
        \centering
        \small
        \setlength{\tabcolsep}{4.0pt}
        \renewcommand{\arraystretch}{1.1}
        \caption{Mean success rates (\%) over three runs on the WindowsAgentArena benchmark. For each model backbone, $\Delta$ denotes the mean EFR result minus the mean POR result in percentage points. The best mean accuracy in each row is bolded; ties are bolded together.}
        \label{tab:windowsagentarena-results}
        \resizebox{\textwidth}{!}{%
        \begin{tabular}{@{}l*{12}{c}@{}}
            \toprule
            \textbf{Task Type} &
            \multicolumn{3}{c}{\textbf{GUI-Owl-32B \citep{ye2025mobileagentv3}}} &
            \multicolumn{3}{c}{\textbf{Qwen3-VL-32B \citep{bai2025qwen3vl}}} &
            \multicolumn{3}{c}{\textbf{Kimi-K2.5 \citep{kimi2025k25blog}}} &
            \multicolumn{3}{c}{\textbf{Seed-1.8 \citep{bytedanceseed2025seed18}}} \\
            \cmidrule(lr){2-4}\cmidrule(lr){5-7}\cmidrule(lr){8-10}\cmidrule(lr){11-13}
            & \textbf{POR} & \textbf{EFR} & \textbf{$\Delta$} & \textbf{POR} & \textbf{EFR} & \textbf{$\Delta$} & \textbf{POR} & \textbf{EFR} & \textbf{$\Delta$} & \textbf{POR} & \textbf{EFR} & \textbf{$\Delta$} \\
            \midrule
            Office & 5.43 & 8.66 & +3.23 & 6.98 & 6.98 & +0.00 & 16.28 & 18.60 & +2.32 & 36.43 & \textbf{37.02} & +0.59 \\
            Web Browsing & 21.33 & 32.76 & +11.43 & 30.00 & 43.22 & +13.22 & 44.11 & \textbf{56.44} & +12.33 & 47.44 & 49.78 & +2.34 \\
            Windows System & 48.61 & 44.44 & -4.17 & 47.22 & 51.39 & +4.17 & 73.61 & 79.17 & +5.56 & \textbf{83.14} & 80.56 & -2.58 \\
            Coding & 58.33 & 75.00 & +16.67 & 66.67 & 65.28 & -1.39 & 73.47 & \textbf{81.81} & +8.34 & 58.33 & 67.92 & +9.59 \\
            Media \& Video & 39.63 & 34.92 & -4.71 & 34.92 & 48.52 & +13.60 & 59.35 & \textbf{61.30} & +1.95 & 24.79 & 26.40 & +1.61 \\
            Windows Utilities & 16.67 & 25.00 & +8.33 & 16.67 & 30.56 & +13.89 & 41.67 & 41.67 & +0.00 & 42.53 & \textbf{58.33} & +15.80 \\
            \midrule
            Overall & 29.04 & 34.24 & +5.20 & 31.60 & 37.55 & +5.95 & 47.40 & \textbf{52.88} & +5.48 & 48.16 & 51.32 & +3.16 \\
            \bottomrule
        \end{tabular}%
        }
    \end{table}
    \begin{figure}[t]
        \centering
        \includegraphics[width=\textwidth]{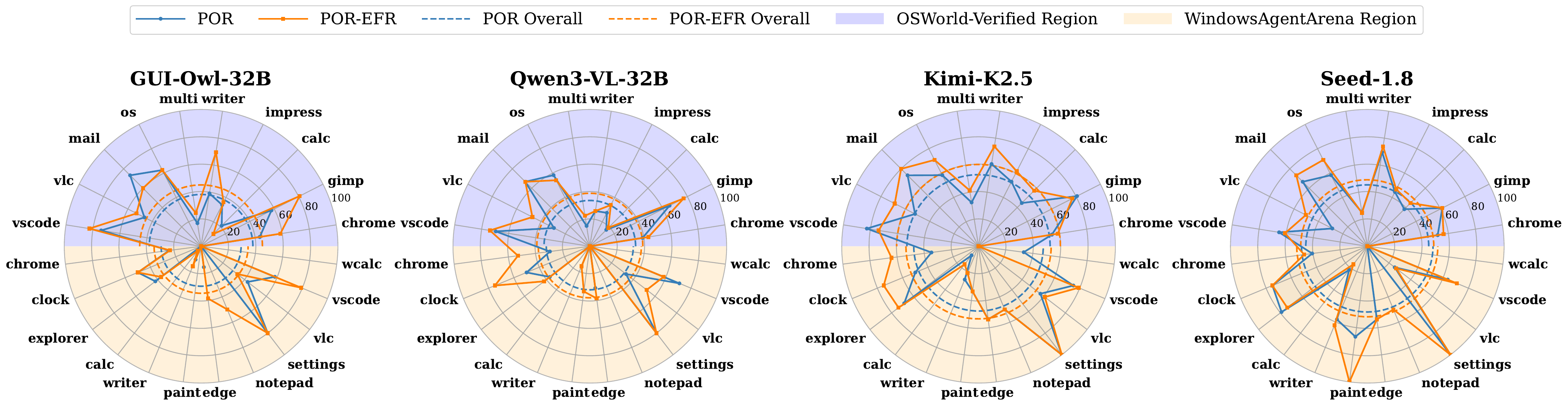}
        \caption{Per-app success-rate comparison between POR and POR-EFR on OSWorld-Verified and WindowsAgentArena. Each subplot corresponds to one backbone model.}
        \label{fig:per-app-radar}
    \end{figure}

    \textbf{Comparison with Baseline.} On OSWorld-Verified, Table~\ref{tab:osworld-results} shows that replacing the standard reflector in POR with EFR consistently improves the mean overall success rate across all four backbones, with an average gain of 5.94\%. The strongest EFR setting, POR-EFR with Kimi-K2.5, achieves the best mean overall score of 59.74\%, improving over its matched POR counterpart by 7.38\%. On WindowsAgentArena, Table~\ref{tab:windowsagentarena-results} similarly shows consistent mean improvements across all four backbones, yielding an average gain of 4.95\%. POR-EFR with Kimi-K2.5 achieves the best mean overall score of 52.88\%, improving over its matched POR counterpart by 5.48\%. Across the three runs, the standard deviations of the overall success rates range from 0.17\% to 1.13\%. Specifically, the POR/EFR standard deviations for GUI-Owl-32B, Qwen3-VL-32B, Kimi-K2.5, and Seed-1.8 are 0.26/0.17, 0.76/0.90, 0.25/0.17, and 0.42/0.56 on OSWorld-Verified, and 0.41/0.63, 0.68/1.13, 0.18/0.25, and 0.54/0.18 on WindowsAgentArena, respectively. These deviations are small relative to the observed overall gains, indicating that the improvements are stable across runs. The corresponding per-app results for the originally reported run are shown in Figure~\ref{fig:per-app-radar}.

    \textbf{Analysis over Different Task Categories.} At the category level, the three-run means show that EFR provides the clearest gains on several task categories rather than uniformly across all of them. On OSWorld-Verified, Kimi-K2.5 improves by 10.23\% and 9.16\% on ``Office'' and ``Workflow'' tasks, reaching mean success rates of 61.20\% and 41.66\%, respectively. On WindowsAgentArena, Kimi-K2.5 improves by 12.33\% on ``Web Browsing'', reaching 56.44\%, while Seed-1.8 improves by 15.80\% on ``Windows Utilities'', reaching 58.33\%. These tasks often involve dense interfaces and frequent state transitions. EFR is particularly beneficial in such settings because it makes page changes more explicit and helps the agent deal with complex situations. We also observe a few category-level decreases compared with the conventional reflector. These cases are scattered across task categories and model backbones, suggesting that they do not indicate a systematic degradation on a specific task type. We provide representative failure examples in the Appendix~\ref{sec:case-study} and summarize their causes in Appendix~\ref{sec:efr-failure-analysis}.

    \textbf{Analysis over Different Model Backbones.} EFR brings consistent mean overall gains across all backbones. On OSWorld-Verified, the improvements are 7.09\%, 5.02\%, 7.38\%, and 4.28\% for GUI-Owl-32B, Qwen3-VL-32B, Kimi-K2.5, and Seed-1.8, respectively. On WindowsAgentArena, the corresponding gains are 5.20\%, 5.95\%, 5.48\%, and 3.16\%. The gains are not determined solely by the capability of the backbone, nor do stronger models make EFR unnecessary. For example, Kimi-K2.5 obtains a larger improvement than GUI-Owl-32B on OSWorld-Verified and a comparable improvement on WindowsAgentArena, indicating that even strong models still benefit from explicit page-difference evidence and serial reflection.

\subsection{Impact of Reflector Judgment on Task Success}
    To analyze the relationship between reflector accuracy and overall task success, we manually annotated the correctness of the reflector’s judgments for each step within the GUI-Owl-32B trajectories on the OSWorld-Verified benchmark. The result is provided in Appendix Figure~\ref{fig:analysis}. Our EFR framework improves the reflector’s judgment accuracy by 7.11\%. A Pearson correlation coefficient of 0.86 indicates a strong positive correlation between reflector accuracy and end-to-end task success across various task categories. These findings suggest that a more reliable reflector enables the agent to evaluate execution outcomes more precisely, thereby enhancing overall performance.

\subsection{Ablation Study}
\label{ablation}
    \begin{table}[t]
        \centering
        \caption{Ablation results of GUI-Owl-32B on OSWorld-Verified.}
        \label{tab:ablation-guiowl-osworld}
        \begin{tabular}{lcc}
            \toprule
            \textbf{Setting} & \textbf{Overall (\%)} & \textbf{Drop (\%)} \\
            \midrule
            POR baseline (w/o both) & 37.86 & -6.97 \\
            Only Two-stage Judging (w/o SoM) & 42.11 & -2.72 \\
            Only SoM (w/o Two-stage Judging) & 38.13 & -6.70 \\
            Diff. Identification w/ Intended Action & 43.45 & -1.38 \\
            Outcome Verification w/o Screenshots & 34.25 & -10.58 \\
            \midrule
            Full method (SoM + Two-stage Judging) & \textbf{44.83} & -- \\
            \bottomrule
        \end{tabular}
    \end{table}
    We conduct an ablation study on the OSWorld-Verified benchmark by using GUI-Owl-32B model.

    \textbf{Effect of SoM.} Removing the SoM module requires the reflector to directly analyze the page differences and subsequently evaluate action success based on global difference descriptions. This configuration results in a 2.72\% decrease in the success rate, demonstrating that SoM facilitates the difference analysis phase by providing localized focus.

    \textbf{Effect of Page Difference Identification.} Ablating the page difference identification module, whereby the reflector evaluates success directly after receiving the marked screenshots results in a significant 6.70\% decrease in the success rate. This drop underscores the critical role of the page difference identification process within our framework. The explicit separation of difference analysis and success judgment is essential for realizing the full benefits of the marked visual evidence.

    \textbf{Replace Grounded with Intended Action in Page Difference Identification Stage.} We further replace the grounded action with the original intended action during the page difference identification stage. This variant leads to a 1.38\% drop in accuracy. Since page difference identification aims to describe what actually changes after execution, conditioning this process on the intended action may cause the VLM to focus on expected outcomes rather than the observed screen transition.

    \textbf{Action Outcome Verification without Screenshots.} We further evaluate a variant in which the action outcome verification stage relies only on the change descriptions, without access to screenshots. This setting results in a 10.58\% drop in accuracy. This degradation suggests that textual change descriptions alone are insufficient for reliable outcome verification in complex GUI environments. When the interface is dense and contains multiple changes, the verifier may perceive only scattered changes, rather than forming a holistic understanding of the page structure.

\subsection{Cost Analysis}
    \begin{figure}[t]
        \centering
        \includegraphics[width=0.32\textwidth]{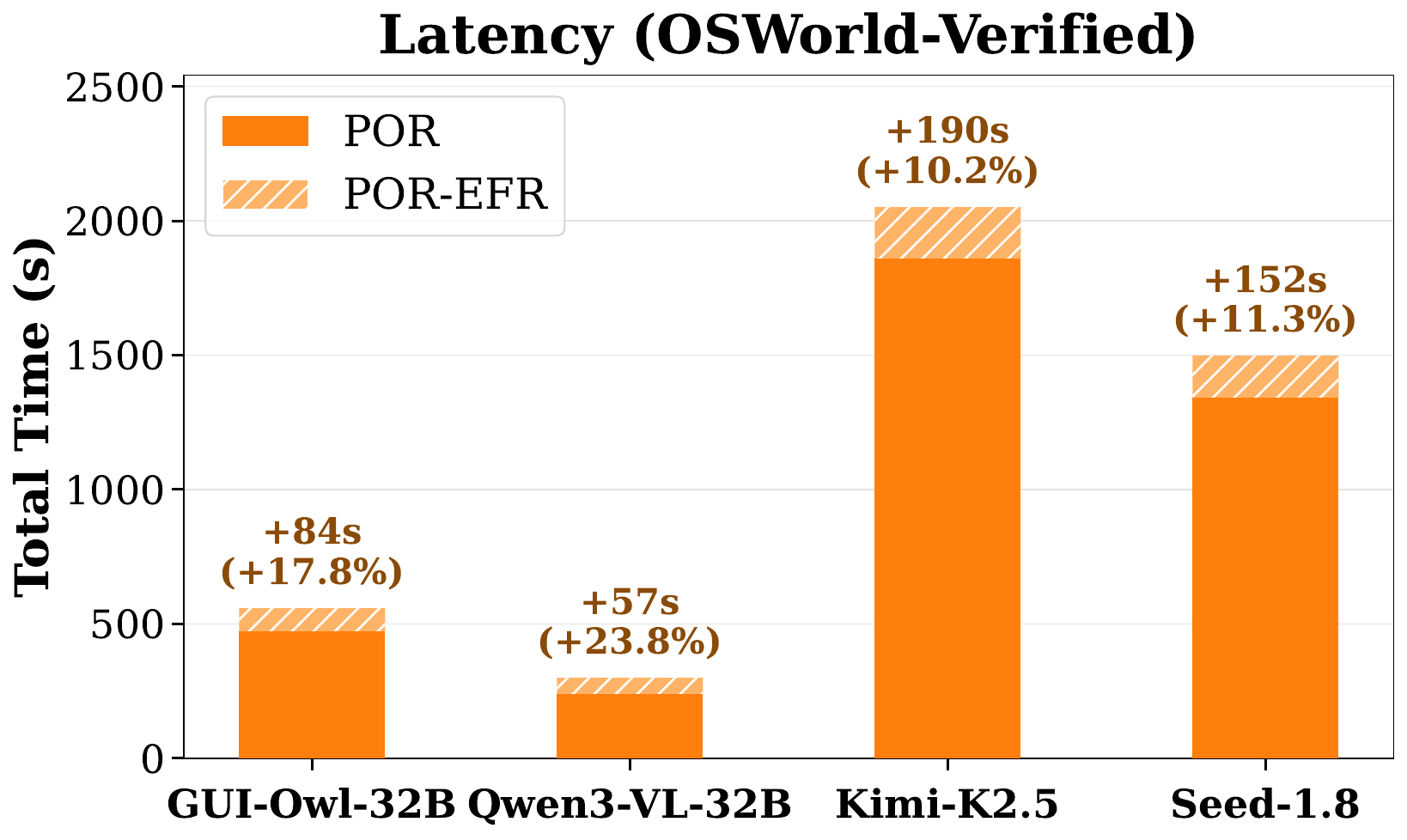}\hfill
        \includegraphics[width=0.32\textwidth]{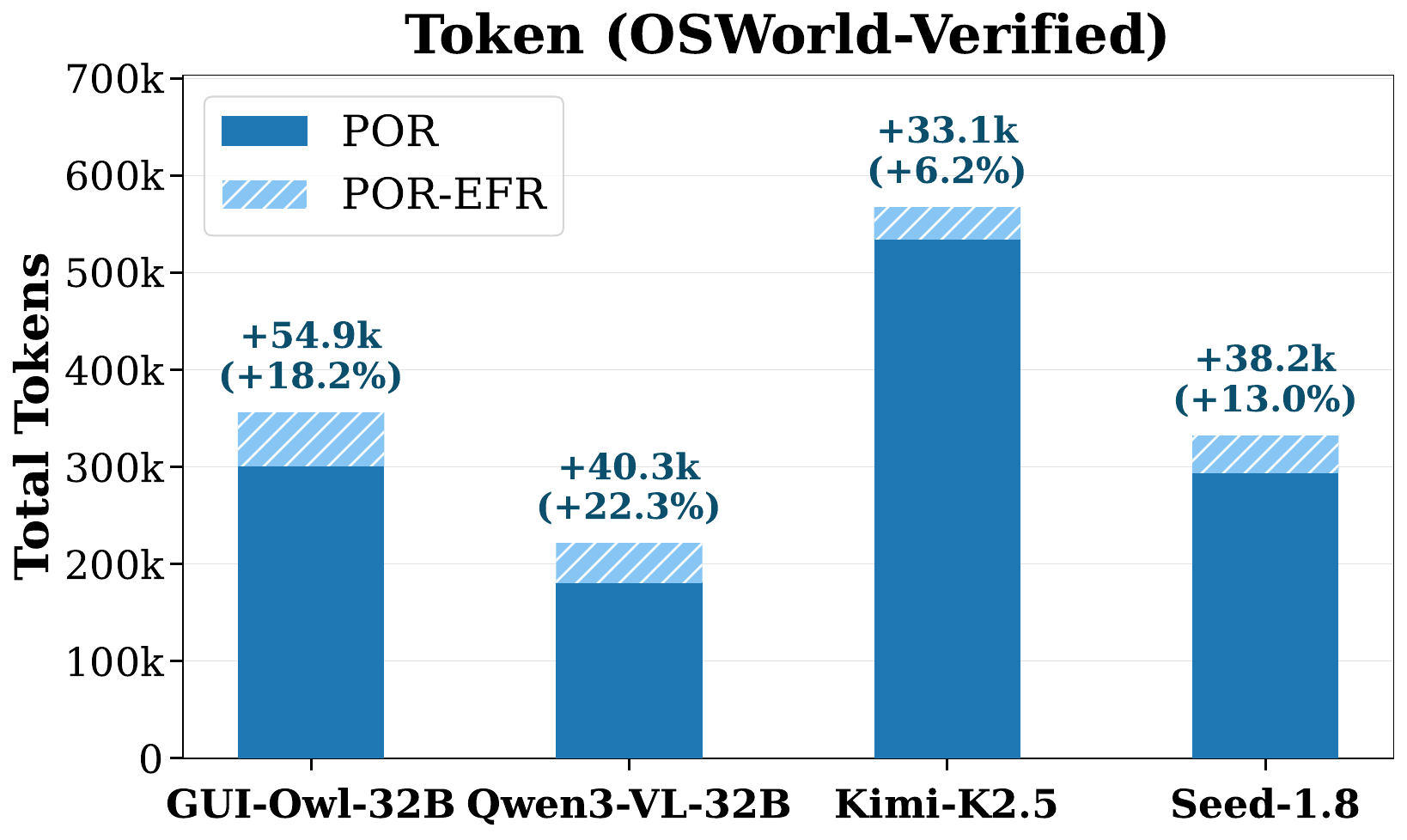}\hfill
        \includegraphics[width=0.32\textwidth]{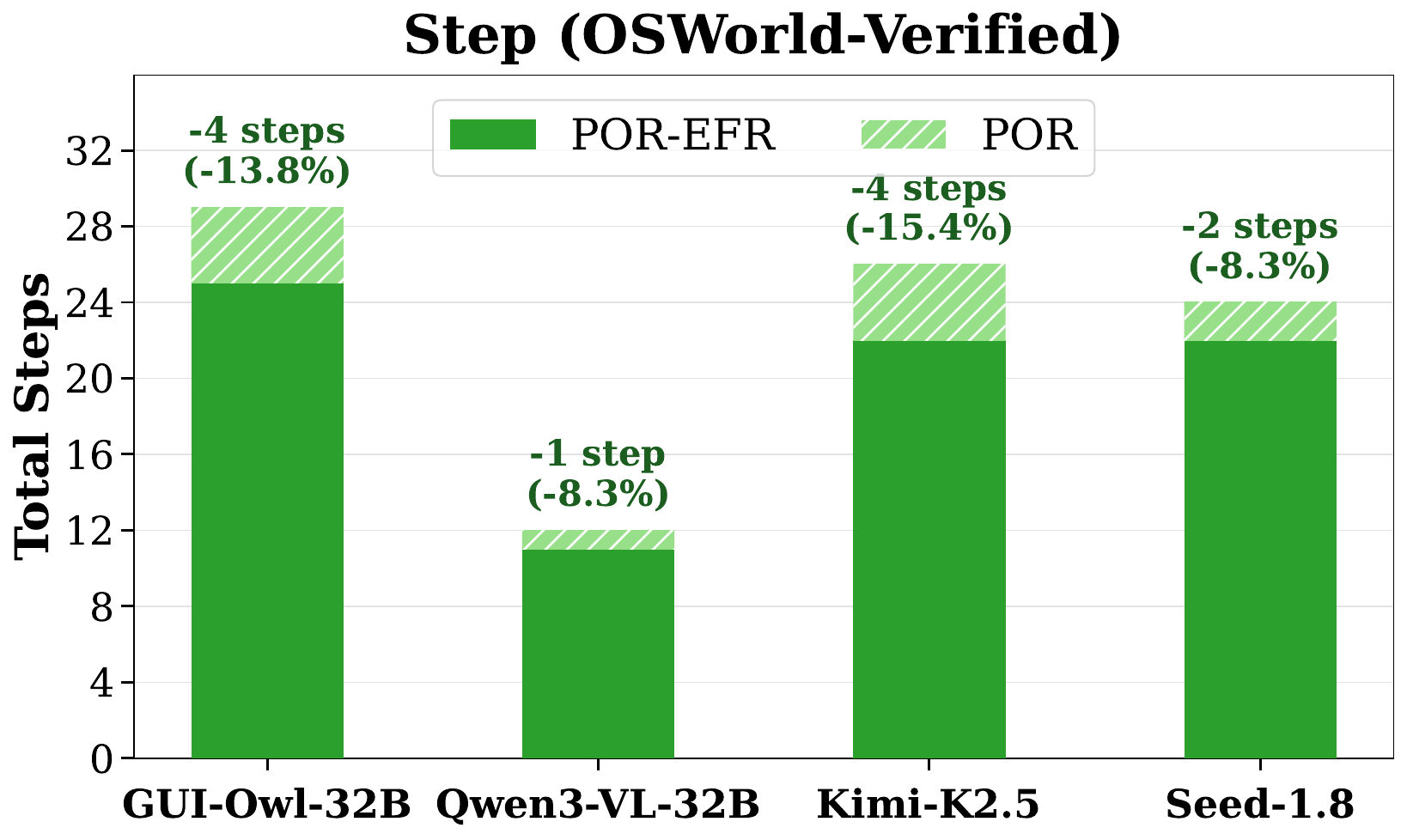}\\[0.6em]

        \includegraphics[width=0.32\textwidth]{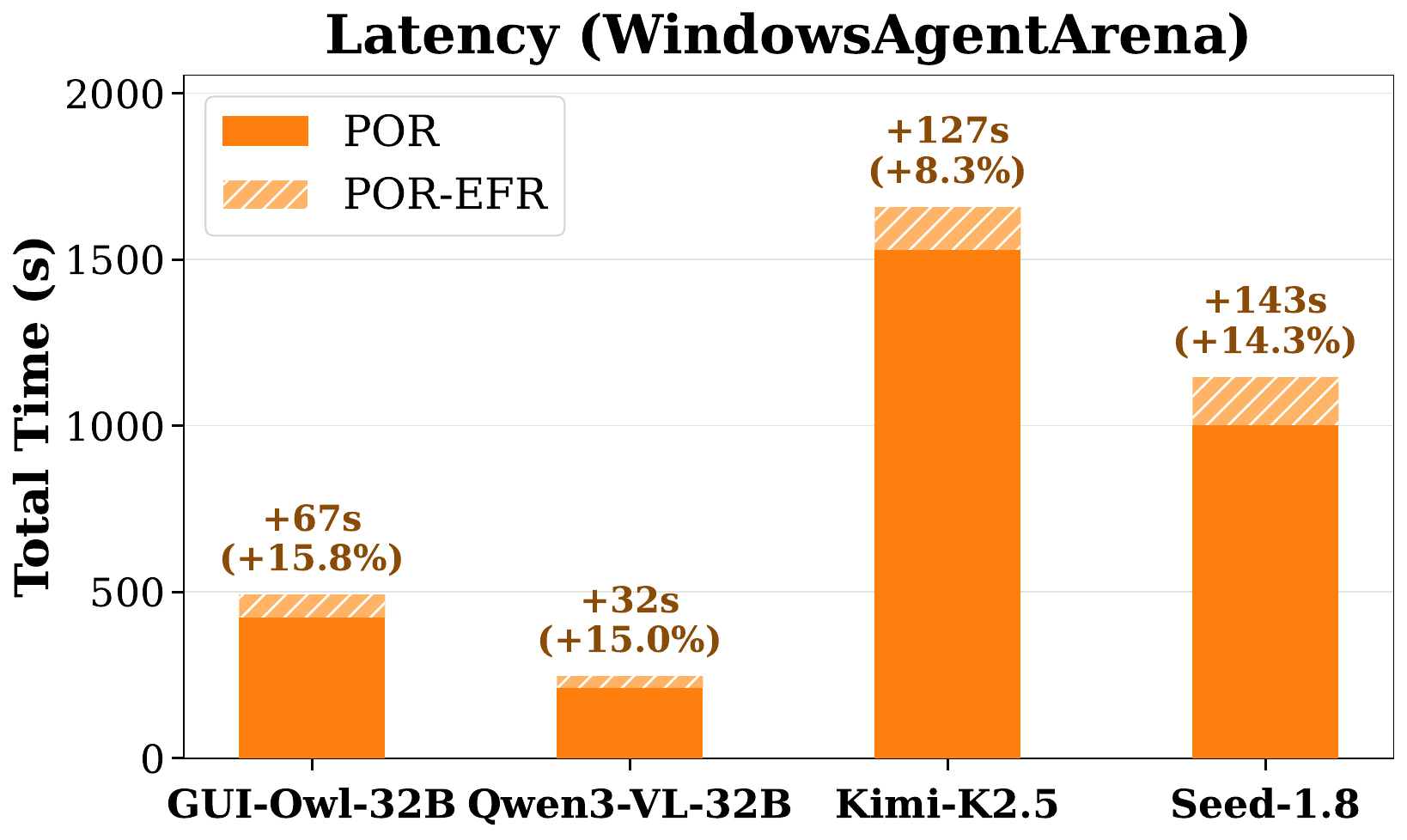}\hfill
        \includegraphics[width=0.32\textwidth]{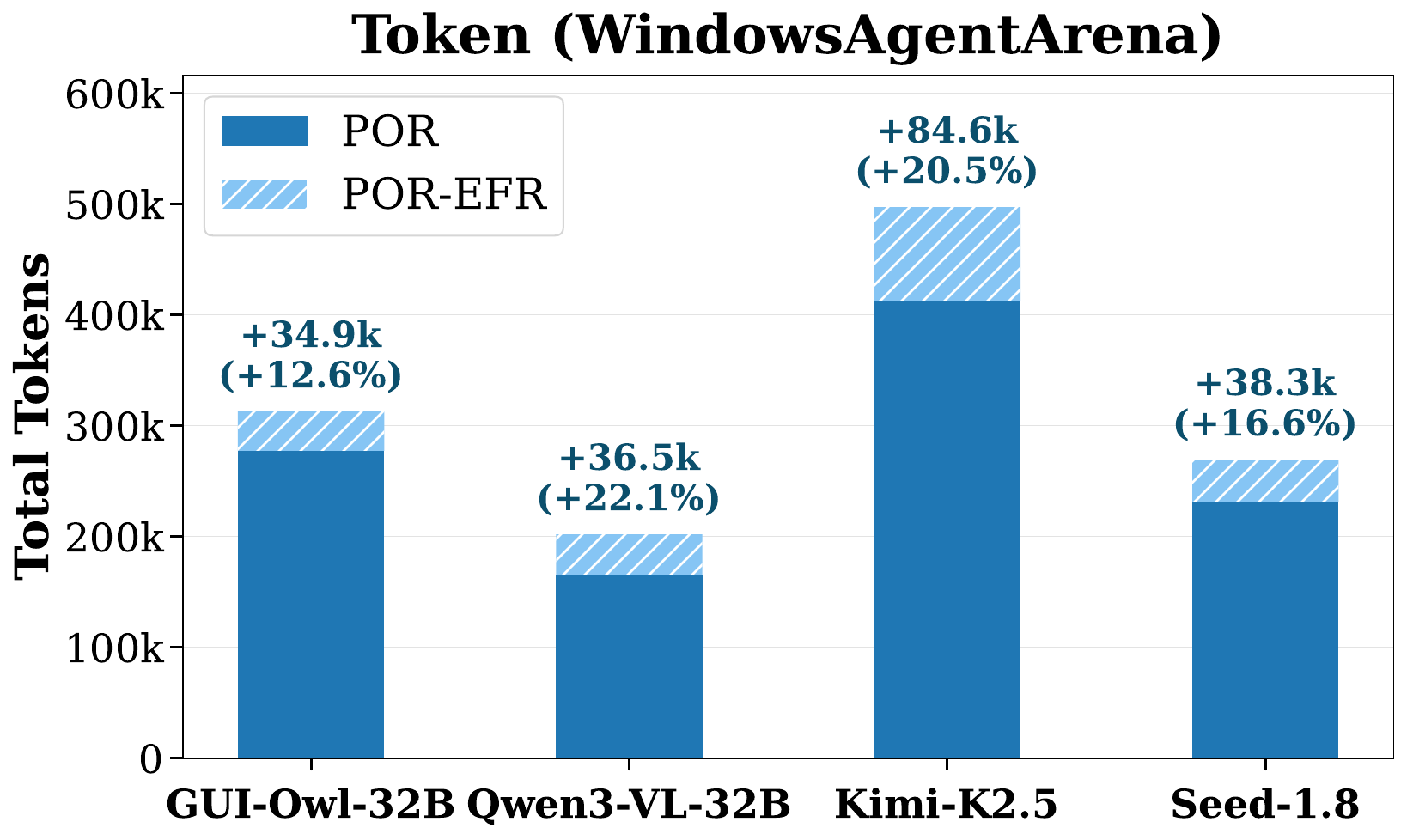}\hfill
        \includegraphics[width=0.32\textwidth]{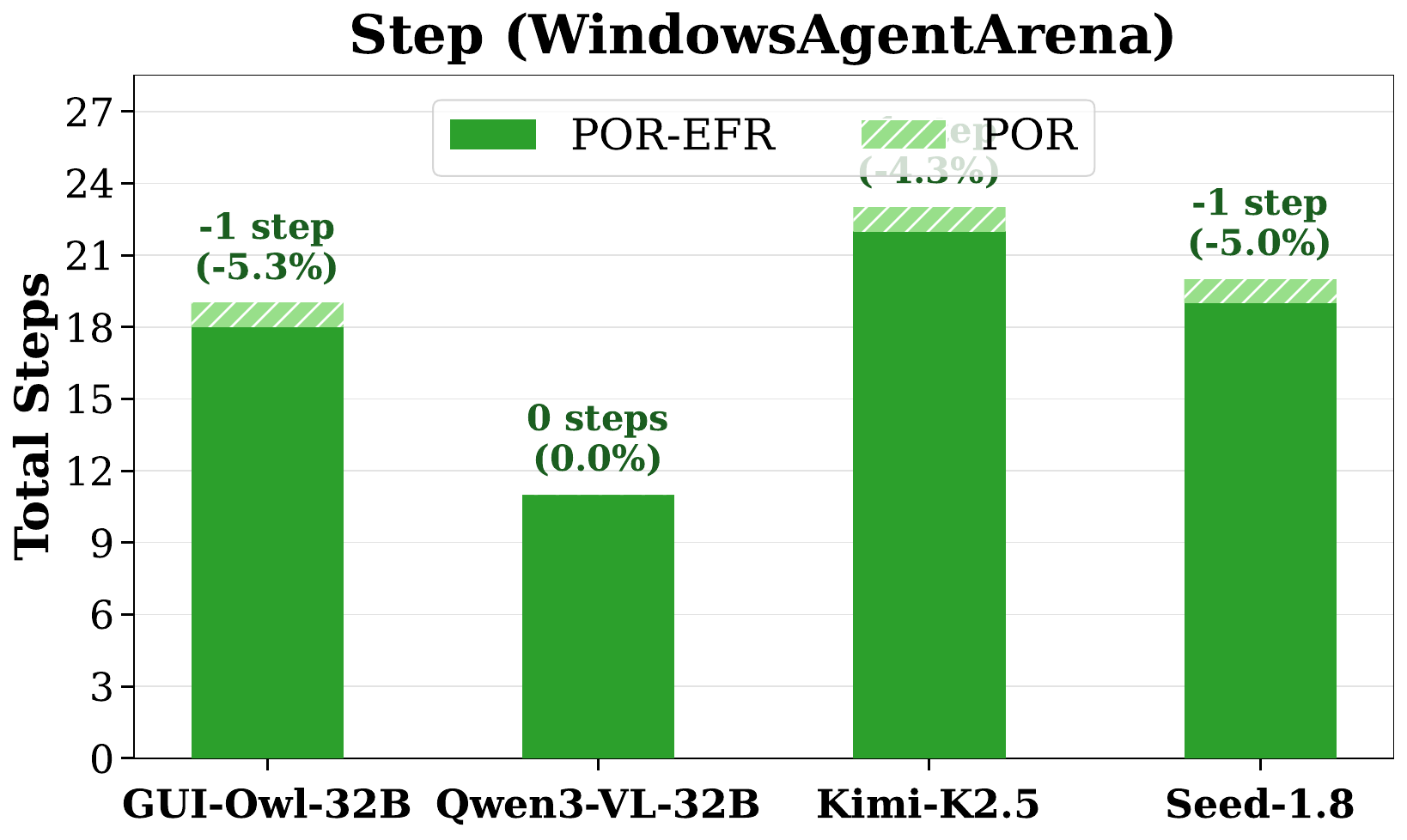}
    
        \caption{Average per-task cost comparison between POR and POR-EFR on OSWorld-Verified and WindowsAgentArena. The first row reports OSWorld-Verified and the second row reports WindowsAgentArena. From left to right, each row shows latency, token usage, and executed steps.}
        \label{fig:cost-analysis}
    \end{figure}

    Figure~\ref{fig:cost-analysis} compares the average per-task cost of POR and POR-EFR on two benchmarks. Averaged over the four backbone models, POR-EFR increases per-task latency by 15.8\% and token usage by 14.9\% on OSWorld-Verified, and by 13.3\% and 17.9\% on WindowsAgentArena, respectively. This overhead mainly stems from the additional VLM call introduced for page-difference analysis, which increases the cost of each reflection step. Although EFR reduces redundant execution by lowering the average number of executed steps by 11.5\% on OSWorld-Verified and 3.7\% on WindowsAgentArena, this reduction only partially offsets the increased per-step reflection cost. Detailed per-step latency and token costs are provided in Appendix~\ref{sec:per-step-cost}.

\section{Conclusion}
    We revisit the role of the reflector in GUI agents and contend that its complexity and functionality are frequently underestimated in desktop environments. We thus introduce EFR, a two-stage reflection framework that decouples explicit visual evidence extraction from subsequent success judgment. Empirical evaluations demonstrate its superior performance. More broadly, our findings highlight that building GUI agents requires more than directly applying existing methods. While POR and SoM provide methodological foundations, their effectiveness depends on how well they are adapted to the perceptual, spatial, and interactive properties of GUI tasks.

\newpage
\bibliographystyle{plainnat}
\bibliography{references}


\newpage
\appendix

\clearpage
\section*{Appendix Contents}
\addcontentsline{toc}{section}{Appendix Contents}
\startcontents[appendix]
\printcontents[appendix]{l}{1}{\setcounter{tocdepth}{2}}
\clearpage

\section{Reflector Accuracy}
\label{sec:reflector}

\begin{figure}[H]
        \centering
        \includegraphics[width=0.8\textwidth]{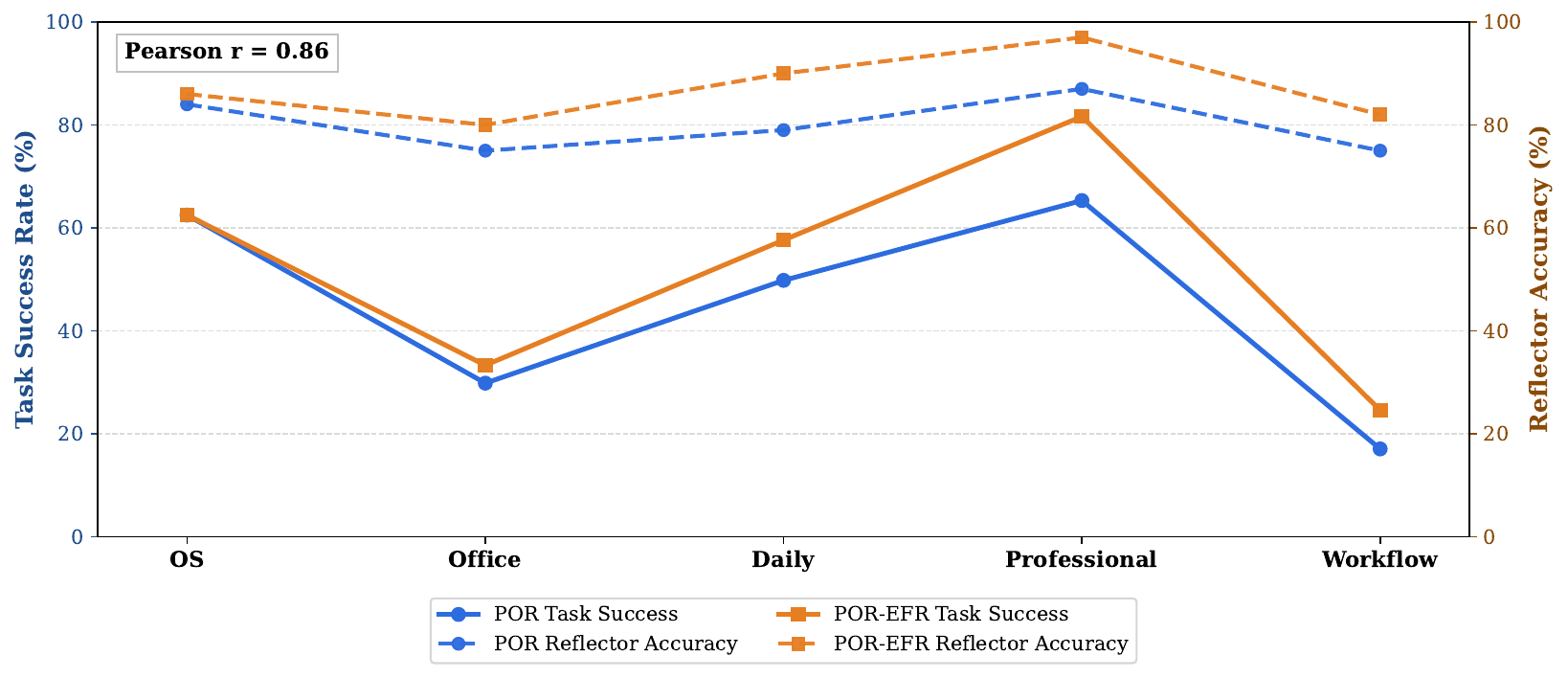}
        \caption{Task success rate vs. reflector success rate in GUI-Owl-32B trajectories on the OSWorld-Verified benchmark.}
        \label{fig:analysis}
    \end{figure}

    We provide two manually annotated analyses based on GUI-Owl-32B trajectories on the OSWorld-Verified benchmark. First, Figure~\ref{fig:analysis} compares reflector judgment accuracy with end-to-end task success across five categories. Overall, replacing the conventional reflector with EFR increases average reflector accuracy from 80.39\% to 87.50\%, while average task success rises from 44.91\% to 51.94\%. The trend is particularly clear on Professional tasks, where reflector accuracy improves by 9.88\% and task success increases by 16.32\%. Across all method-category points in the figure, reflector accuracy and task success have a Pearson correlation of 0.86, suggesting that more reliable intermediate verification is strongly associated with higher final task completion.

    Second, Table~\ref{tab:appendix-reflector-failure-causes} reports how often failed trajectories can be attributed to reflector errors under the conventional POR framework and EFR. EFR reduces reflector-attributed failures from 61 of 223 failed trajectories (27.35\%) to 27 of 199 (13.57\%), a decrease of 13.78\%. More specifically, page-difference identification errors decrease from 43 of 223 failures (19.28\%) to 14 of 199 (7.04\%), a 12.25\% reduction, while success-judgment errors decrease from 18 of 223 (8.07\%) to 13 of 199 (6.53\%), a 1.54\% reduction. Thus, EFR reduces both error types, with the larger improvement arising from more reliable page-difference identification.

    \begin{table}[H]
        \centering
        \small
        \setlength{\tabcolsep}{4.5pt}
        \renewcommand{\arraystretch}{1.1}
        \caption{Manual failure attribution for GUI-Owl-32B under the conventional POR framework and EFR on OSWorld-Verified, using the same model and mutually exclusive error categories. ``Diff. Error'' denotes page-difference identification errors, ``Judge. Error'' denotes success-judgment errors given correctly identified page differences, and ``Total'' denotes all failed trajectories. For the reflector-error rate, $\Delta$ denotes EFR minus POR in percentage points. The lower rate in each row is bolded; ties are bolded together.}
        \label{tab:appendix-reflector-failure-causes}
        \resizebox{\textwidth}{!}{%
        \begin{tabular}{@{}l*{11}{c}@{}}
            \toprule
            \textbf{Application} &
            \multicolumn{2}{c}{\textbf{Reflector}} &
            \multicolumn{2}{c}{\textbf{Diff. Error}} &
            \multicolumn{2}{c}{\textbf{Judge. Error}} &
            \multicolumn{2}{c}{\textbf{Total}} &
            \multicolumn{3}{c}{\textbf{Rate (\%)}} \\
            \cmidrule(lr){2-3}\cmidrule(lr){4-5}\cmidrule(lr){6-7}\cmidrule(lr){8-9}\cmidrule(lr){10-12}
            & \textbf{POR} & \textbf{EFR}
            & \textbf{POR} & \textbf{EFR}
            & \textbf{POR} & \textbf{EFR}
            & \textbf{POR} & \textbf{EFR}
            & \textbf{POR} & \textbf{EFR} & \textbf{$\Delta$} \\
            \midrule
            chrome & 3 & 0 & 2 & 0 & 1 & 0 & 26 & 19 & 11.54 & \textbf{0.00} & -11.54 \\
            gimp & 0 & 0 & 0 & 0 & 0 & 0 & 11 & 5 & \textbf{0.00} & \textbf{0.00} & +0.00 \\
            libreoffice\_calc & 23 & 11 & 16 & 7 & 7 & 4 & 37 & 41 & 62.16 & \textbf{26.83} & -35.33 \\
            libreoffice\_impress & 12 & 2 & 8 & 0 & 4 & 2 & 31 & 30 & 38.71 & \textbf{6.67} & -32.04 \\
            libreoffice\_writer & 4 & 1 & 4 & 1 & 0 & 0 & 14 & 7 & 28.57 & \textbf{14.29} & -14.28 \\
            multi\_apps & 17 & 11 & 12 & 6 & 5 & 5 & 77 & 70 & 22.08 & \textbf{15.71} & -6.37 \\
            os & 0 & 0 & 0 & 0 & 0 & 0 & 9 & 9 & \textbf{0.00} & \textbf{0.00} & +0.00 \\
            thunderbird & 0 & 0 & 0 & 0 & 0 & 0 & 4 & 6 & \textbf{0.00} & \textbf{0.00} & +0.00 \\
            vlc & 1 & 1 & 1 & 0 & 0 & 1 & 9 & 8 & \textbf{11.11} & 12.50 & +1.39 \\
            vs\_code & 1 & 1 & 0 & 0 & 1 & 1 & 5 & 4 & \textbf{20.00} & 25.00 & +5.00 \\
            \midrule
            \textbf{All} & 61 & 27 & 43 & 14 & 18 & 13 & 223 & 199 & 27.35 & \textbf{13.57} & -13.78 \\
            \bottomrule
        \end{tabular}%
        }
    \end{table}

\section{Step-Level Reflection Example}
\label{sec:step-level-reflection-example}

\begin{figure}[t]
        \centering
        \begin{subfigure}[t]{0.48\textwidth}
            \centering
            \includegraphics[width=\linewidth]{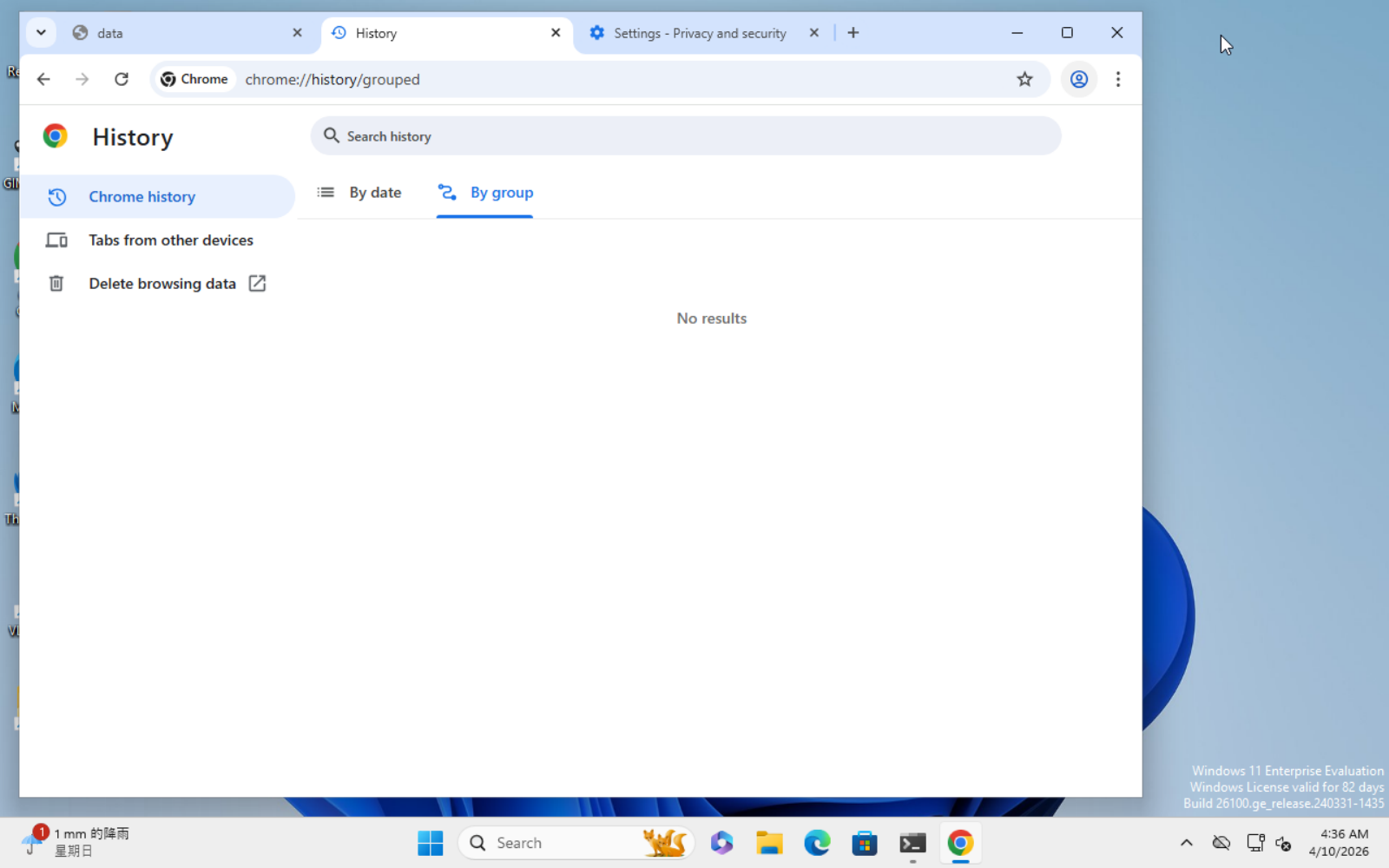}
            \caption{Raw pre-action screenshot}
            \label{fig:step-example-pre-state}
        \end{subfigure}\hfill
        \begin{subfigure}[t]{0.48\textwidth}
            \centering
            \includegraphics[width=\linewidth]{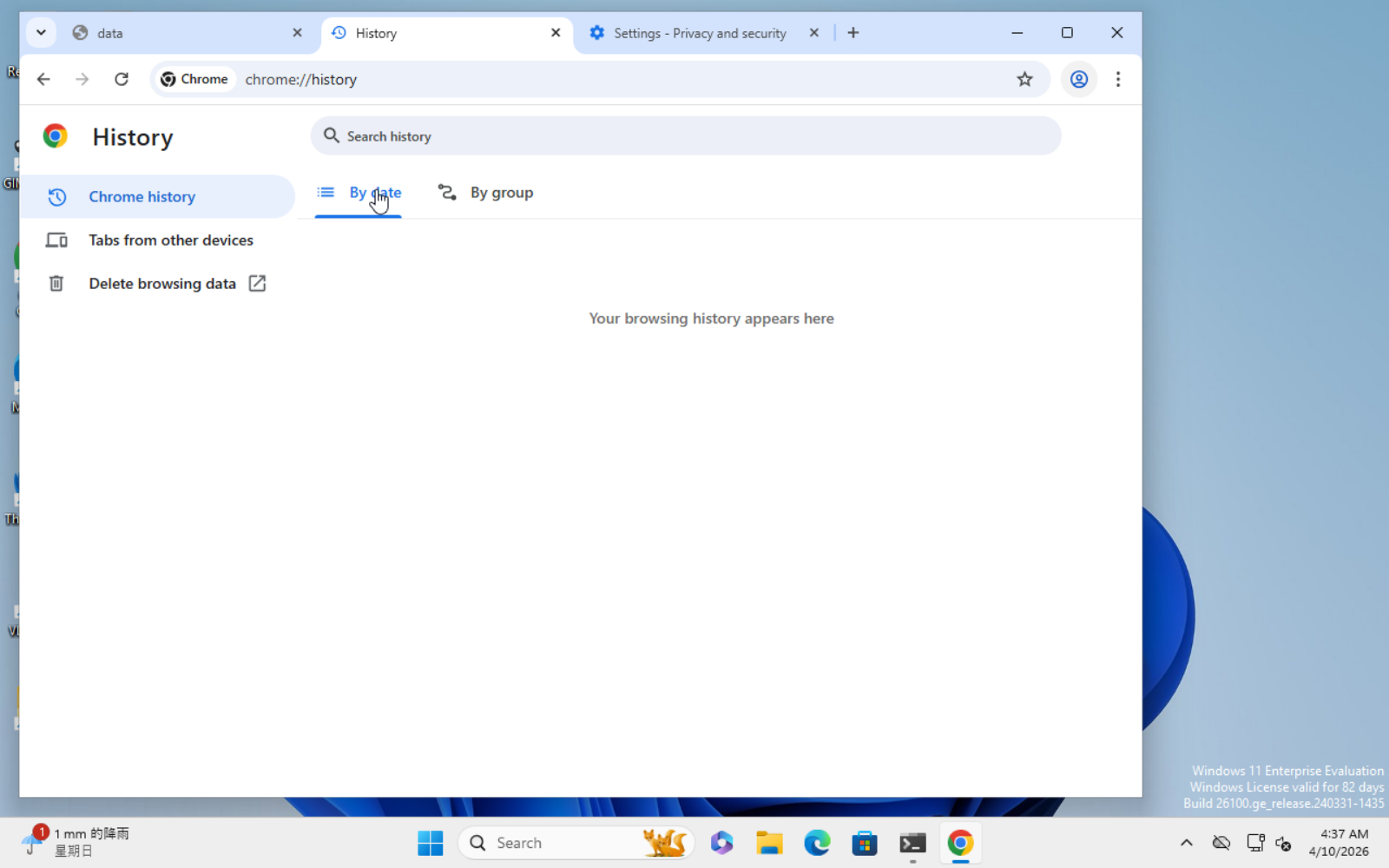}
            \caption{Raw post-action screenshot}
            \label{fig:step-example-post-state}
        \end{subfigure}

        \vspace{0.45em}
        \begin{subfigure}[t]{0.48\textwidth}
            \centering
            \includegraphics[width=\linewidth]{figures/som_example/pre_trace.pdf}
            \caption{SoM-marked pre-action screenshot}
            \label{fig:step-example-pre-trace}
        \end{subfigure}\hfill
        \begin{subfigure}[t]{0.48\textwidth}
            \centering
            \includegraphics[width=\linewidth]{figures/som_example/post_trace.pdf}
            \caption{SoM-marked post-action screenshot}
            \label{fig:step-example-post-trace}
        \end{subfigure}

        \vspace{0.45em}
        \begin{subfigure}[t]{0.48\textwidth}
            \centering
            \includegraphics[width=\linewidth]{figures/som_example/before_marked.pdf}
            \caption{Clean evidence before action}
            \label{fig:step-example-before-marked}
        \end{subfigure}\hfill
        \begin{subfigure}[t]{0.48\textwidth}
            \centering
            \includegraphics[width=\linewidth]{figures/som_example/after_marked.pdf}
            \caption{Clean evidence after action}
            \label{fig:step-example-after-marked}
        \end{subfigure}

        \caption{A one-step reflection example for clicking the ``By date'' button in Chrome History. The clean evidence keeps Areas 1--4 and removes Area 5, which only reflects the taskbar clock update.}
        \label{fig:appendix-step-reflection-example}
    \end{figure}

    We provide a concrete one-step example to illustrate the difference between the conventional reflector and EFR. The intended action is \texttt{Click the ``By date'' button in Chrome History to sort the history view by date}, while the actually executed grounded action is \texttt{Click at coordinate (519,232)}. This is a simple example in which both reflectors make the correct judgment. The raw screenshots and EFR intermediate screenshots are shown in Figure~\ref{fig:appendix-step-reflection-example}. The conventional reflector directly reasons over the raw pre-action and post-action screenshots, whereas EFR generates explicit intermediate evidence before making the final reflection judgment.

    \subsection{Conventional Reflector}
    \begin{itemize}[leftmargin=*, labelsep=0.5em]
        \item \textbf{Step 1: One-step action verification.}
        
        \textbf{VLM input:} the intended action: \texttt{Click the ``By date'' button in Chrome History to sort the history view by date}. The raw pre-action and post-action screenshots: Figure~\ref{fig:step-example-pre-state} and Figure~\ref{fig:step-example-post-state}.
        
        \textbf{VLM output:} The observed screen changes from the grouped history view to the date-based history view, indicate that the action succeeded.
    \end{itemize}

    \subsection{EFR}
    \begin{itemize}[leftmargin=*, labelsep=0.5em]
        \item \textbf{Step 1: SoM annotation.}

        \textbf{SoM module input:} The raw pre-action and post-action screenshots: Figure~\ref{fig:step-example-pre-state} and Figure~\ref{fig:step-example-post-state}. The grounded action: \texttt{Click at coordinate (519,232)}.

        \textbf{SoM module output:} The marked pre-action screenshot: Figure~\ref{fig:step-example-pre-trace}, page change regions and action position are marked. The marked post-action screenshot: Figure~\ref{fig:step-example-post-trace}, only page change regions are marked.

        \item \textbf{Step 2: Page-difference identification.}

        \textbf{VLM input:} the SoM-marked pair in Figures~\ref{fig:step-example-pre-trace} and \ref{fig:step-example-post-trace}, plus the grounded action \texttt{Click at coordinate (519,232)}.

        \textbf{VLM output:} a structured evidence description: Area 1 shows that the mouse cursor moves away; Area 2 shows that the URL changes from \texttt{chrome://history/grouped} to \texttt{chrome://history}; Area 3 shows that the selected history mode changes from ``By group'' to ``By date''; Area 4 shows that the center message changes from ``No results'' to ``Your browsing history appears here''; Area 5 is identified as environment noise because it only reflects the taskbar clock update. The clean evidence output is Figure~\ref{fig:step-example-before-marked} and Figure~\ref{fig:step-example-after-marked}, which retain Areas 1--4 and remove Area 5.

        \item \textbf{Step 3: Evidence-based action verification.}

        \textbf{VLM input:} the intended action \texttt{Click the ``By date'' button in Chrome History to sort the history view by date}, the structured evidence from Step 2, and the clean evidence screenshots in Figures~\ref{fig:step-example-before-marked} and \ref{fig:step-example-after-marked}.

        \textbf{Actual output:} The action succeeds because Areas 2--4 show the page changing from the grouped history view to the date-based history view, which matches the intended action.
    \end{itemize}

\section{Set-of-Marks Annotation Details}
\label{sec:som-detail}

    \subsection{Visual Change Area Annotation}

        Algorithm~\ref{alg:diff-annotation} summarizes the visual difference annotation procedure used in EFR. The algorithm takes a pair of pre- and post-action screenshots and outputs two consistently annotated images together with the merged difference boxes. At this stage, the goal is only to identify \emph{where} visual changes occur; the action-location annotation step is described in the next subsection. All visual-change markers use red bounding boxes, with each box assigned a unique index so that the reflector can refer to the same changed region across the pre- and post-action screenshots.

        \par\smallskip
        \begingroup
        \small
        \noindent\rule{\linewidth}{0.5pt}
        \par
        \refstepcounter{algorithm}\label{alg:diff-annotation}
        \textbf{Algorithm \thealgorithm}: Visual Difference Region Annotation
        \par
        \noindent\rule{\linewidth}{0.5pt}
        \begin{algorithmic}[1]
            \Require Pre-action screenshot $S_{pre}$, post-action screenshot $S_{post}$
            \Ensure Annotated screenshots $\hat{S}_{pre}, \hat{S}_{post}$ and merged box set $\mathcal{B}$
            \Function{ShouldMerge}{$b_i, b_j, M$}
                \State Compute the union box $u$ of $b_i$ and $b_j$
                \State Compute the $x$/$y$ gaps between $b_i$ and $b_j$
                \State Compute the difference density of $u$ on binary mask $M$
                \If{$b_i$ and $b_j$ are tightly adjacent}
                    \State \Return density$(u) \geq \tau_{\mathrm{dens}}$
                \Else
                    \State \Return False
                \EndIf
            \EndFunction
            \State Convert $S_{pre}$ and $S_{post}$ to grayscale images
            \State $D \gets$ pixel-wise absolute difference between the two grayscale images
            \State $M \gets$ binary mask obtained by thresholding $D$
            \State Extract external contours from $M$ and convert each contour to a bounding box
            \State Remove boxes with a size of at most $2$ pixels
            \State Remove boxes fully contained in the top status-bar area
            \State Expand each remaining box by $5$ pixels on all sides
            \State $\mathcal{B} \gets$ the resulting box set
            \Repeat
                \State changed $\gets$ False
                \ForAll{pairs $(b_i, b_j)$ in $\mathcal{B}$}
                    \If{\Call{ShouldMerge}{$b_i, b_j, M$}}
                        \State Replace $(b_i, b_j)$ with their union in $\mathcal{B}$
                        \State changed $\gets$ True
                        \State \textbf{break}
                    \EndIf
                \EndFor
            \Until{changed = False}
            \State Sort $\mathcal{B}$ from top to bottom and from left to right
            \State Draw the same red bounding boxes and indices in $\mathcal{B}$ on $S_{pre}$ and $S_{post}$
            \State \Return annotated images $\hat{S}_{pre}, \hat{S}_{post}$ and $\mathcal{B}$
        \end{algorithmic}
        \noindent\rule{\linewidth}{0.5pt}
        \par
        \endgroup
        \smallskip

        The output of Algorithm~\ref{alg:diff-annotation} constitutes the candidate visual evidence used by the reflector. However, these boxes alone do not indicate where the current action was actually executed. To connect the detected visual changes with the concrete operation site, we further annotate the action location on the pre-action screenshot in the next subsection.

    \subsection{Action Location Annotation}

        Compared with the difference-region identification in Algorithm~\ref{alg:diff-annotation}, action-location annotation is a lighter-weight post-processing step. Based on the geometric parameters associated with the actual action $A_{ground}$ executed by the operator, it directly visualizes where the action takes place on the annotated pre-action screenshot $\hat{S}_{pre}$, thereby providing the reflector with a spatial anchor aligned with the observed visual changes. We denote this process as
        \begin{equation}
            (\tilde{S}_{pre}, L) = f_{\mathrm{act}}(\hat{S}_{pre}, A_{ground}).
            \label{eq:act-annotation}
        \end{equation}
        where $\tilde{S}_{pre}$ denotes the pre-action screenshot annotated with the action marker, and $L$ denotes the geometric location information of the action.

        In our implementation, actions are divided into three categories in this stage, and the detailed action types are provided in Appendix~\ref{sec:action-space}. For actions that can be reduced to a single screen point, we directly read the execution coordinates $(x, y)$ and map them onto the pixel grid. We then draw a blue hollow circle centered at that location on $S_{pre}$ to highlight the actual operation site. For the \texttt{drag} action, we further parse the drag start point $(x_s, y_s)$ and end point $(x_e, y_e)$ from the executable code. The drag trajectory is visualized with yellow markers: the start point is shown as a yellow hollow circle, the end point is shown as a yellow filled circle, and the two points are connected by a yellow arrow indicating the movement direction. For actions that do not require explicit screen coordinates, such as keyboard shortcuts or waiting actions, no action-location annotation is added. Accordingly, the geometric representations of point-based actions, drag actions, and coordinate-free actions are written as
        \begin{equation}
            L_{\mathrm{point}} = \bigl((x, y)\bigr), \qquad
            L_{\mathrm{drag}} = \bigl((x_s, y_s), (x_e, y_e)\bigr), \qquad
            L_{\mathrm{none}} = \varnothing.
            \label{eq:act-geometry}
        \end{equation}

        This action annotation step does not involve any additional visual inference; it merely projects the parameters of the action already executed by the operator back into the screenshot space. Its purpose is to explicitly indicate where the actual operation took place, thereby assisting the reflector in its observation. Finally, the action-location marker and the candidate difference boxes together form $S_{pre}^{som}$ in the main text, while the post-action screenshot retains only the difference boxes to form $S_{post}^{som}$.

\section{Benchmark Task Grouping Details}
\label{sec:benchmark-grouping-details}

    We summarize the correspondence between the app-level categories and the domain-level categories used in Tables~\ref{tab:osworld-results} and \ref{tab:windowsagentarena-results}. The left table reports OSWorld-Verified and the right table reports WindowsAgentArena. The category mappings follow the official benchmark definitions \citep{xie2024osworld,bonatti2025windowsagentarena}.
    \begin{center}
        \captionsetup{type=table}
        \centering
        \scriptsize
        \setlength{\tabcolsep}{3pt}
        \renewcommand{\arraystretch}{1.1}
        \captionof{table}{Correspondence between the original app-level partition and the merged task categories used in the main results.}
        \label{tab:benchmark-grouping-mapping}
        \begin{subtable}[t]{0.48\textwidth}
            \centering
            \caption{OSWorld-Verified category mapping.}
            \label{tab:osworld-merged-breakdown}
            \begin{tabular}{@{}>{\raggedright\arraybackslash}p{0.22\linewidth}>{\raggedright\arraybackslash}p{0.46\linewidth}>{\raggedleft\arraybackslash}p{0.12\linewidth}>{\raggedleft\arraybackslash}p{0.12\linewidth}@{}}
                \toprule
                \textbf{Domain} & \textbf{Included Apps} & \textbf{\# Tasks} & \textbf{\%} \\
                \midrule
                OS & OS tasks & 24 & 6.6\% \\
                Office & LibreOffice Calc, Impress, Writer & 117 & 32.4\% \\
                Daily & Chrome, Thunderbird, VLC & 78 & 21.6\% \\
                Professional & VS Code, GIMP & 49 & 13.6\% \\
                Workflow & Tasks involving multiple apps & 93 & 25.8\% \\
                \midrule
                Total & -- & 361 & 100.0\% \\
                \bottomrule
            \end{tabular}
        \end{subtable}\hfill
        \begin{subtable}[t]{0.48\textwidth}
            \centering
            \caption{WindowsAgentArena category mapping.}
            \label{tab:waa-merged-breakdown}
            \begin{tabular}{@{}>{\raggedright\arraybackslash}p{0.22\linewidth}>{\raggedright\arraybackslash}p{0.46\linewidth}>{\raggedleft\arraybackslash}p{0.12\linewidth}>{\raggedleft\arraybackslash}p{0.12\linewidth}@{}}
                \toprule
                \textbf{Domain} & \textbf{Included Apps} & \textbf{\# Tasks} & \textbf{\%} \\
                \midrule
                Office & LibreOffice Calc, LibreOffice Writer & 43 & 27.9\% \\
                Web Browsing & Chrome, Microsoft Edge & 30 & 19.5\% \\
                Windows System & File Explorer, Settings & 24 & 15.6\% \\
                Coding & VS Code & 24 & 15.6\% \\
                Media \& Video & VLC & 21 & 13.6\% \\
                Windows Utilities & Clock, Paint, Notepad, Windows Calculator & 12 & 7.8\% \\
                \midrule
                Total & -- & 154 & 100.0\% \\
                \bottomrule
            \end{tabular}
        \end{subtable}
    \end{center}

\section{Complete EFR Prompt}
\label{sec:reflector-prompt}

    We report the prompt of the two-stage EFR reflector. To make the runtime assembly explicit, we replace concrete values with bracketed placeholders.

    \textbf{Stage 1: Page Difference Identification: }
    \begin{tcolorbox}[breakable,colback=orange!5!white, colframe=orange!50!black, title=Prompt in stage 1 (System prompt):]
        You are a GUI Visual Change Analyzer. Your task is to identify and describe page changes that are directly related to a specific action, based on the provided action description and screenshots.

        \textbf{INPUT DATA}:
        
        1.  Action Description: Text explaining the user's intended operation.
        
        2.  Before-Action Screenshot (first attached image): Contains Action Markers (e.g., blue circles, yellow arrows) indicating the operation location. It also contains the same Red Bounding Boxes with unique IDs as the After Image, highlighting the regions where visual differences were detected between the two screenshots.
        
        3.  After-Action Screenshot (second attached image): Contains Red Bounding Boxes with unique IDs highlighting all detected visual differences. The area outside the Red Bounding Boxes has not undergone any changes.
        
        4.  Bounding Boxes List: Contains the list of the coordinates of the Red Bounding Boxes, mapped to Red Box IDs in both images.
        
        \textbf{OUTPUT STRUCTURE:}
        
        Step 1: Filtering Analysis
        
        [List every Box ID and judge it. Format: ID - Reason - Verdict]
        
        * Area 1: [It is just the system clock changing time. Not relevant to the action.] - [Noise]
        
        * Area 2: [It is a dropdown menu appearing near the click location, which is caused by the click action.] - [Relevant]

        * Area 3: [It is the system clock changing time and a menu appearing near the click location, caused by the click action.] - [Relevant]
        
        Step 2: Final Description
        
        [Based ONLY on the ``Relevant'' areas from Step 1, generate the final description.]
        
        1. Detailed Changes: [Describe specific changes inside relevant red boxes by logical groups.]
        
          - Areas [List of Relevant IDs]: [E.g., ``- Areas [2, 5]: A dropdown menu appeared showing options 'Save' and 'Exit'.'']
        
        2. Description: [Summarize the visual changes caused by the action in a concise manner, referencing only the Relevant areas. E.g., ``After clicking 'File', a dropdown menu appeared at the top-left corner showing options 'Save' and 'Exit'.'']
    \end{tcolorbox}

    \begin{tcolorbox}[colback=blue!5!white, colframe=blue!50!black, title=Prompt in stage 1 (User prompt):]
        \textbf{Input Data:}
        
        Action Description: []
        
        Bounding Box List (mapped to Red Box IDs in images): []
    \end{tcolorbox}

    \textbf{Stage 2: Action Outcome Verification: }
    \begin{tcolorbox}[breakable,colback=blue!5!white, colframe=blue!50!black, title=Prompt in stage 2 (System prompt):]
        You are a GUI Task Verifier. Your role is to determine if the previous action was successful.

        \textbf{INPUT DATA}:
        
        1. Before-Action Screenshot (first attached image): This image shows the GUI state BEFORE the action was executed. Red Bounding Boxes with IDs have been artificially overlaid on this image to mark WHERE changes will happen.
        
        2. After-Action Screenshot (second attached image): This image shows the GUI state AFTER the action was executed. The same Red Bounding Boxes with IDs are overlaid to highlight the regions where changes occurred.
        
        3. Visual Change Description (provided in text): This is the ABSOLUTE GROUND TRUTH text report describing what actually changed AFTER the action. You MUST prioritize and trust the ``Visual Change Description''.
        
        \textbf{Your Task:}
        Compare the before and after screenshots along with the ``Visual Change Description'' against the ``Latest Action'' to decide if the action achieved its intended goal.
    \end{tcolorbox}

    \begin{tcolorbox}[breakable,colback=blue!5!white, colframe=blue!50!black, title=Prompt in stage 2 (User prompt):]
        \textbf{Latest Action}
        
        []
        
        \textbf{Visual Changes After Action}
        
        []
        
        ---
        Carefully examine the information provided above to determine whether the last action produced the expected behavior. If the action was successful, update the progress status accordingly. If the action failed, identify the failure mode and provide reasoning on the potential reason causing this failure.
        
        Pro Tip: In rare cases, the UI might not visibly change even if a click action is performed correctly — for example, when clicking on a color before drawing. In such situations, you can assume the action was successful and proceed — for example, by drawing a line.
        When the user instruction involves adjusting some values (e.g., brightness, contrast, steps), be sure to check if the values meet expectations after the operation. If the adjusted value is very close to the target, treat the action as successful and note the remaining gap in Progress Status.
        
        Provide your output in the following format containing four parts:
        
        \textbf{Analyse}
        (Describe the criteria for judgment)
        
        Final Visual State:[Describe the state AFTER the action by comparing the before screenshot (first image) and after screenshot (second image), and cross-referencing with the ``Visual Changes After Action'' text. Remember to absolutely trust the text if there's a conflict. Then describe if the final visual state meets the expected status or makes reasonable progress towards it]
        
        \textbf{Outcome}
        
        Choose from the following options. Give your response as ``A'', ``B'' or ``C'':
        
        A: Successful or Partially Successful. The result of the last action meets the expectation, OR the action was successfully executed and made incremental progress towards the goal (even if the final objective is not completely fulfilled yet).
        
        B: Failed. The reason for the failure is a grounding error.
        
        C: Failed. The reason for the failure is an error in the action itself.
        
        \textbf{Error Description}
        
        If the action failed (Outcome B or C), provide a detailed description of the error and the potential reason causing this failure. If the action succeeded or partially succeeded (Outcome A), put ``None'' here.
        
        \textbf{Progress Status}
        
        Update the progress status.
    \end{tcolorbox}
    
\section{Action Space}
\label{sec:action-space}

    To remain consistent with the underlying execution interface and to avoid introducing additional confounding factors into the reflector analysis due to differences in low-level action API design, we adopt the action space definition of Mobile-Agent-v3 \citep{ye2025mobileagentv3}. Specifically, the operator first outputs a structured action. If the action contains an element description, a grounding model is first invoked to resolve the corresponding element into screen coordinates, after which the grounded result is compiled into executable desktop-operation code. Table~\ref{tab:action-space} summarizes the action types used in this work, their semantic meanings, and the corresponding executable code templates. For readability, we omit the shared \texttt{import pyautogui} prefix in the table and retain only the core execution statements.

    \begin{table}[H]
        \centering
        \small
        \renewcommand{\arraystretch}{1.15}
        \caption{Action space directly adopted from Mobile-Agent-v3 \citep{ye2025mobileagentv3}. The code column shows the executable template after grounding. Shared import prefixes are omitted for brevity.}
        \label{tab:action-space}
        \begin{tabular}{p{0.16\textwidth}p{0.24\textwidth}p{0.50\textwidth}}
            \toprule
            \textbf{Action Type} & \multicolumn{2}{c}{\textbf{Action Details}} \\
            \cmidrule(lr){2-3}
            & \textbf{Description} & \textbf{Executable Code Template} \\
            \midrule
            \texttt{click} & Click at a grounded screen location. & \texttt{pyautogui.click(x=x, y=y)} \\
            \texttt{double\_click} & Double-click at a grounded screen location. & \texttt{pyautogui.doubleClick(x=x, y=y)} \\
            \texttt{right\_click} & Right-click at a grounded screen location. & \texttt{pyautogui.rightClick(x=x, y=y)} \\
            \texttt{type} & Focus an input region and type text, optionally clearing the field or pressing Enter. & \parbox[t]{0.50\textwidth}{\raggedright \texttt{pyautogui.doubleClick(x=x, y=y);} \newline \texttt{hotkey(``ctrl'', ``a''); press(``delete'');} \newline \texttt{pyautogui.typewrite(text, interval=1.0);} \newline \texttt{pyautogui.press(``enter'')}} \\
            \texttt{hotkey} & Execute a keyboard shortcut combination. & \texttt{pyautogui.hotkey(key1, ..., keyn)} \\
            \texttt{scroll} & Move to a grounded location and scroll by a signed amount. & \texttt{pyautogui.moveTo(x, y); pyautogui.scroll(value)} \\
            \texttt{wait} & Pause execution for a specified duration. & \texttt{time.sleep(t)} \\
            \texttt{done} & Mark the current task as completed. & \texttt{DONE} \\
            \texttt{drag} & Drag from one grounded location to another grounded location. & \parbox[t]{0.50\textwidth}{\raggedright \texttt{pyautogui.moveTo(x1, y1);} \newline \texttt{pyautogui.dragTo(x2, y2, duration=1.0);} \newline \texttt{pyautogui.mouseUp()}} \\
            \bottomrule
        \end{tabular}
    \end{table}

    In terms of grounding, \texttt{click}, \texttt{double\_click}, \texttt{right\_click}, \texttt{type}, and \texttt{scroll} each require resolving only a single screen point; \texttt{drag} requires separately resolving both the start point and the end point; whereas \texttt{hotkey}, \texttt{wait}, and \texttt{done} do not depend on visual grounding. This also explains why the action-location annotation in the previous subsection only needs to distinguish between three cases.

\section{Per-Step Cost Analysis}
\label{sec:per-step-cost}

\begin{figure}[H]
        \centering
        \includegraphics[width=0.48\textwidth]{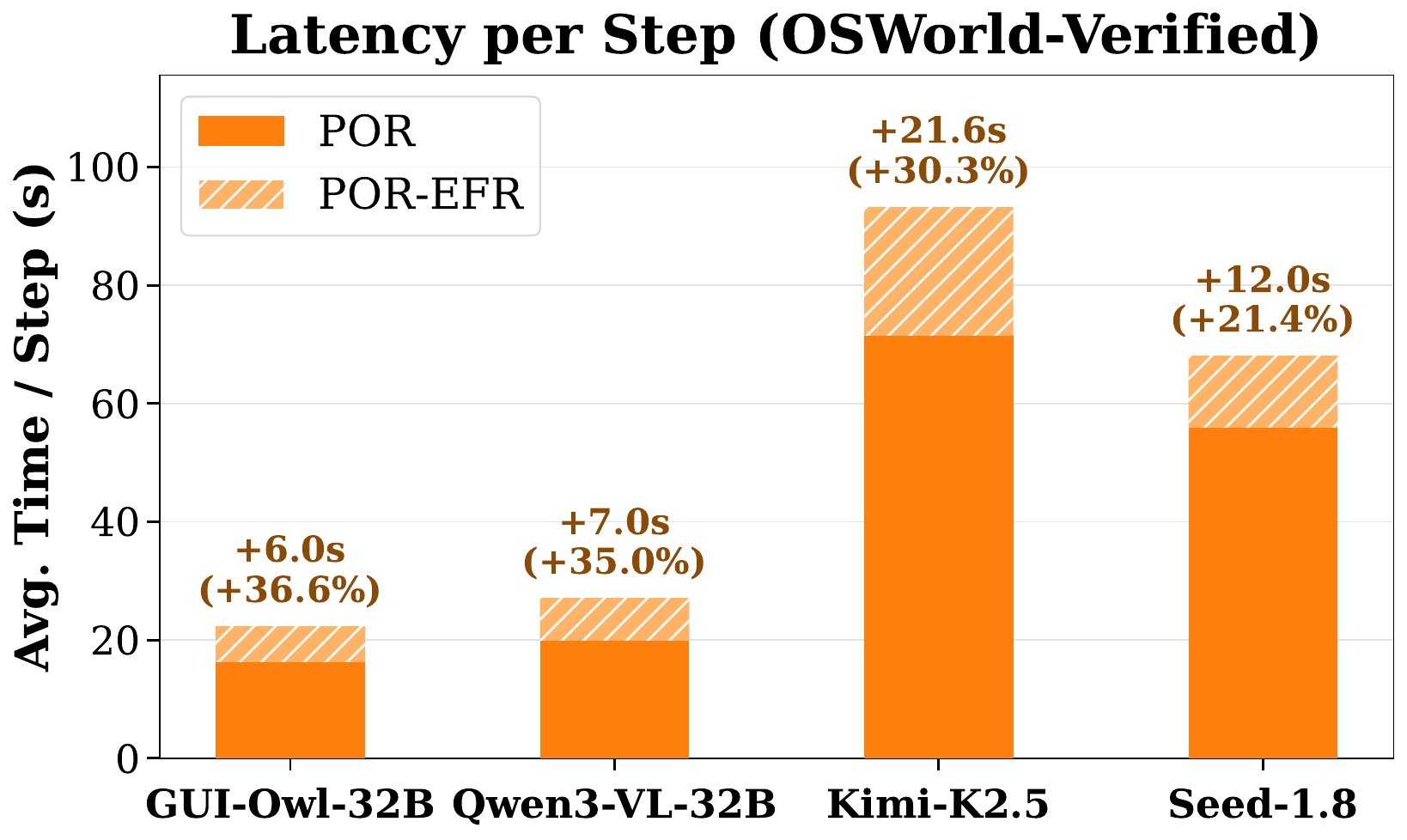}\hfill
        \includegraphics[width=0.48\textwidth]{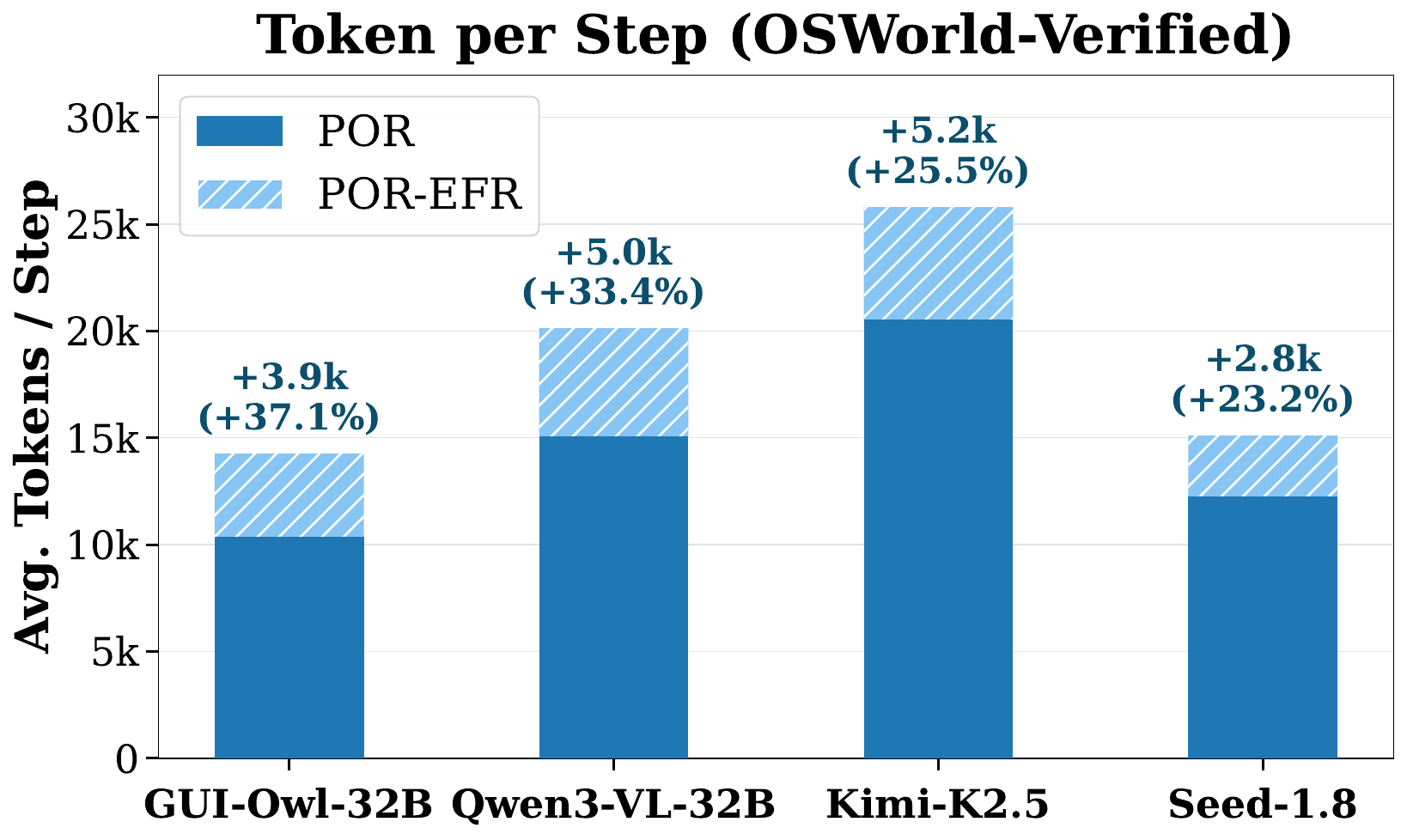}\\[0.6em]
        \includegraphics[width=0.48\textwidth]{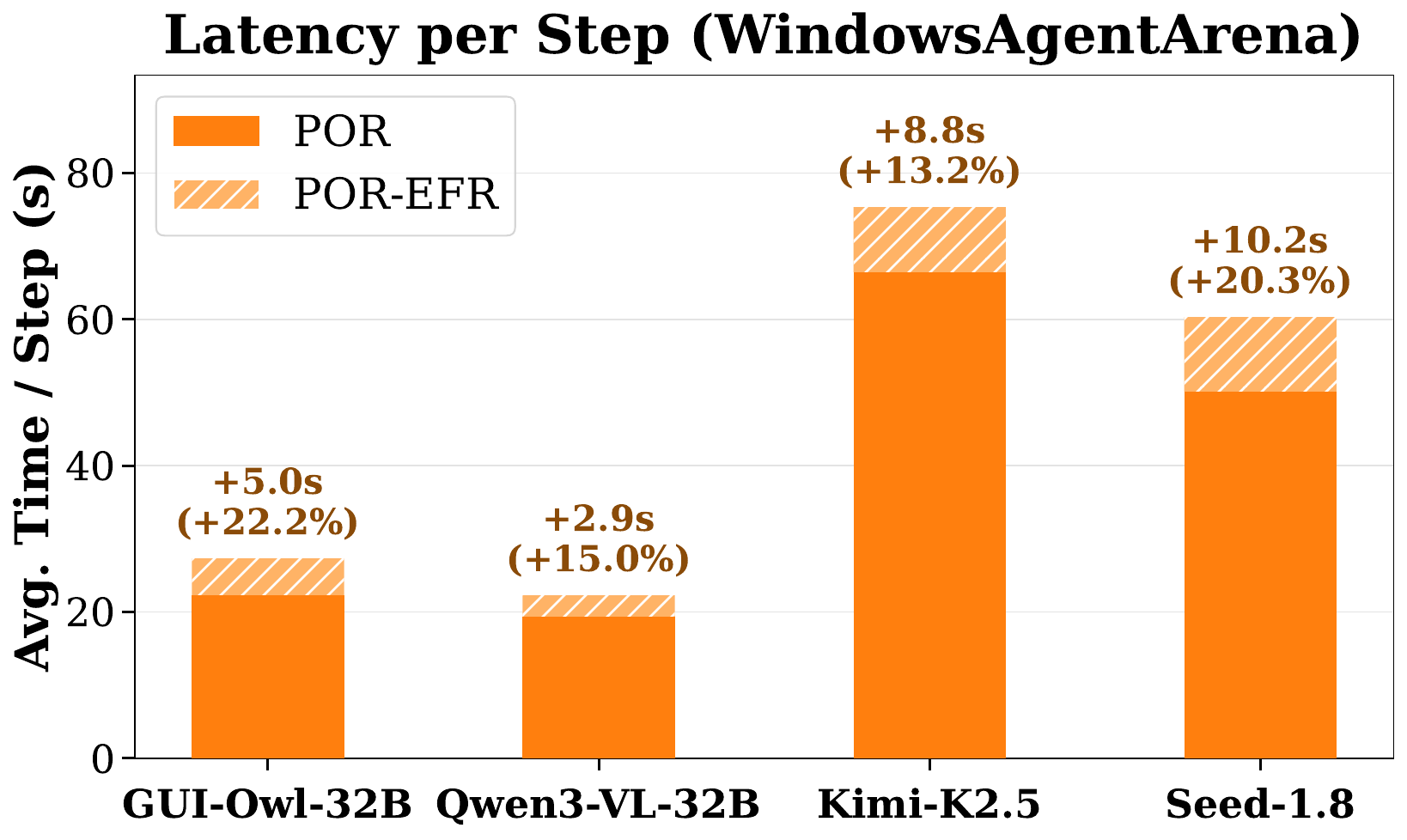}\hfill
        \includegraphics[width=0.48\textwidth]{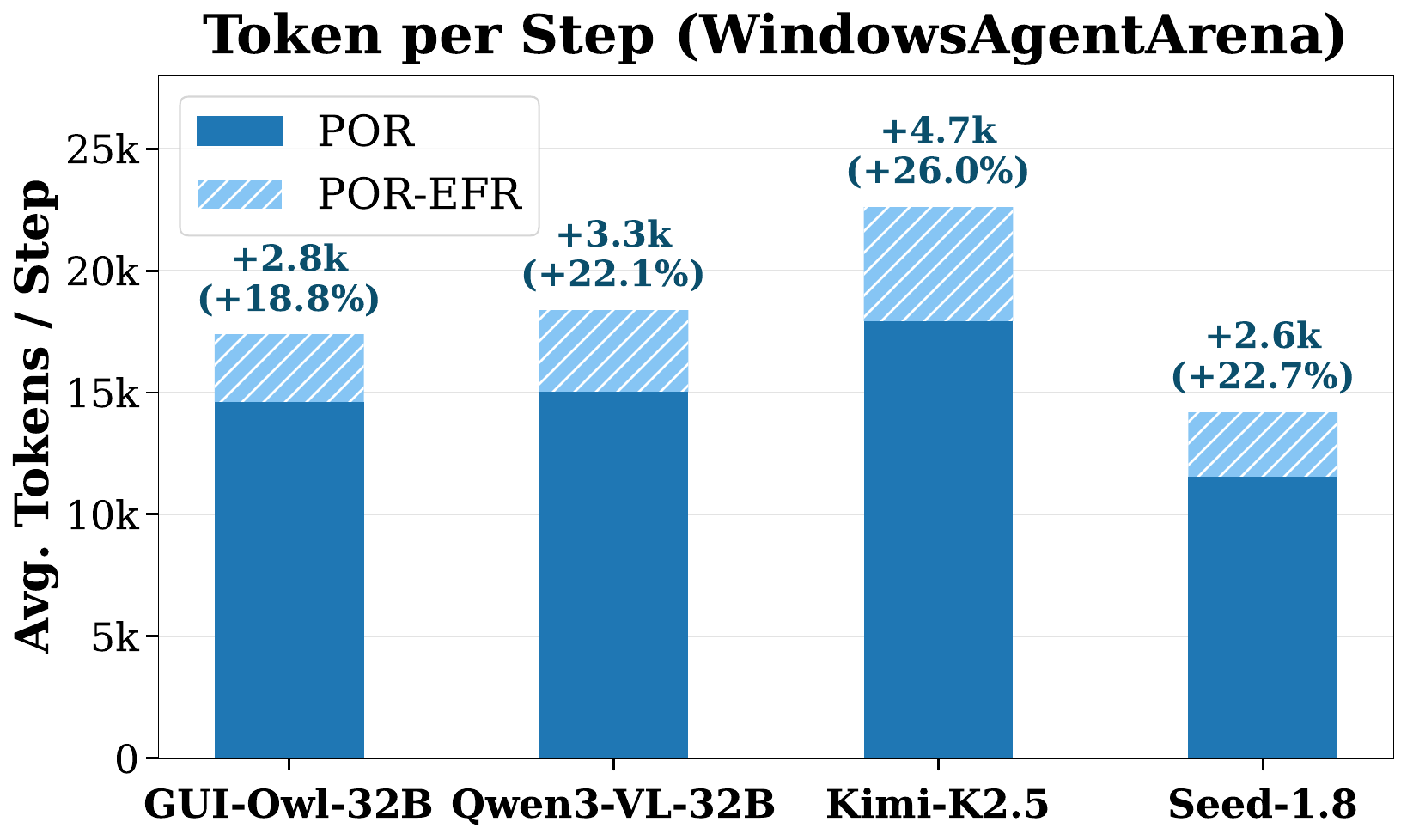}
        \caption{Average per-step cost comparison between POR and POR-EFR on OSWorld-Verified and WindowsAgentArena. The first row reports OSWorld-Verified and the second row reports WindowsAgentArena. From left to right, each row shows latency and token usage averaged over executed steps.}
        \label{fig:per-step-cost-analysis}
    \end{figure}

    Figure~\ref{fig:per-step-cost-analysis} reports the average latency and token usage per executed step. POR-EFR increases latency and token usage per step by 30.8\% and 29.8\% on OSWorld-Verified, and by 17.7\% and 22.4\% on WindowsAgentArena, respectively. These results show that EFR introduces additional cost at each reflection step, while the step reduction shown in Figure~\ref{fig:cost-analysis} helps keep the final per-task overhead moderate.

\section{Case Study}
\label{sec:case-study}

\begin{figure}[H]
        \centering
        \includegraphics[width=\textwidth]{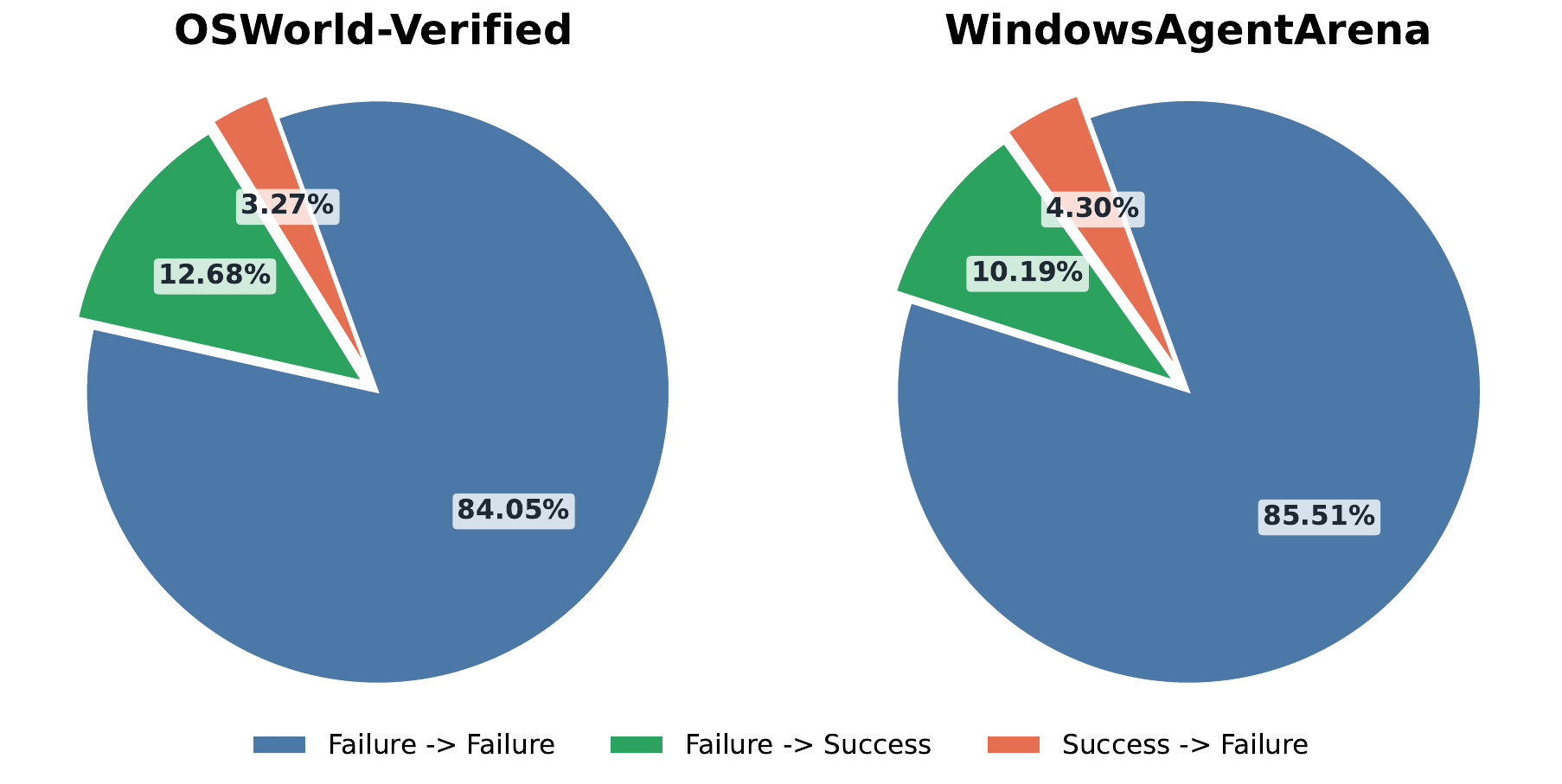}
        \caption{Outcome transition distributions on OSWorld-Verified and WindowsAgentArena. Each pie chart shows the proportion of trajectories that remain failed, are corrected from failure to success, or regress from success to failure after applying our method.}
        \label{fig:pie}
    \end{figure}

    Before examining individual trajectories, we first summarize how EFR changes task-level outcomes relative to the conventional reflector, as illustrated in Figure~\ref{fig:pie}. On OSWorld-Verified, 84.05\% of samples remain failure$\rightarrow$failure, 12.68\% are converted from failure$\rightarrow$success, and 3.27\% shift from success$\rightarrow$failure. On WindowsAgentArena, the corresponding proportions are 85.51\%, 10.19\%, and 4.30\%, respectively. These transition statistics show that EFR improves task success mainly by recovering previously failed samples while introducing only a small fraction of negative flips. At the same time, the large failure$\rightarrow$failure proportion indicates that many GUI-agent tasks remain unsolved even after applying our method, highlighting that robust GUI agents still require substantial further progress. We illustrate each of these three types with representative examples.

    \subsection{Failure to Success}
    \label{sec:fail-to-success}

    In this section, we compare two paired trajectories from OSWorld-Verified. Each pair contains a successful trajectory with EFR and a failed trajectory with the original one-step reflector. Case~1 illustrates the ``small-change'' regime, where the decisive evidence is only a tiny cursor or formatting change. Case~2 illustrates the ``distributed-change'' regime, where the target cell looks correct but the action also damages unrelated regions. In both cases, the operator can produce similar low-level actions; the key difference is whether the reflector can first organize the relevant visual evidence before deciding success.

    \subsubsection{Case 1: Change the 2 in ``H2O'' to a subscript in LibreOffice Writer}

    This case corresponds to \path{libreoffice_writer/0b17a146-2934-46c7-8727-73ff6b6483e8}. The task is to convert the ``2'' in ``H2O'' into a subscript. The success and failure trajectories differ mainly in how the reflector treats very subtle page changes. This example shows how EFR improves the reflector in the small-change regime: by explicitly grounding judgment in observed evidence, it converts a nearly invisible cursor change into a usable signal, preventing repeated false failure judgments.

    \begin{center}
        \begin{minipage}{\textwidth}
            \centering
            \includegraphics[width=\linewidth]{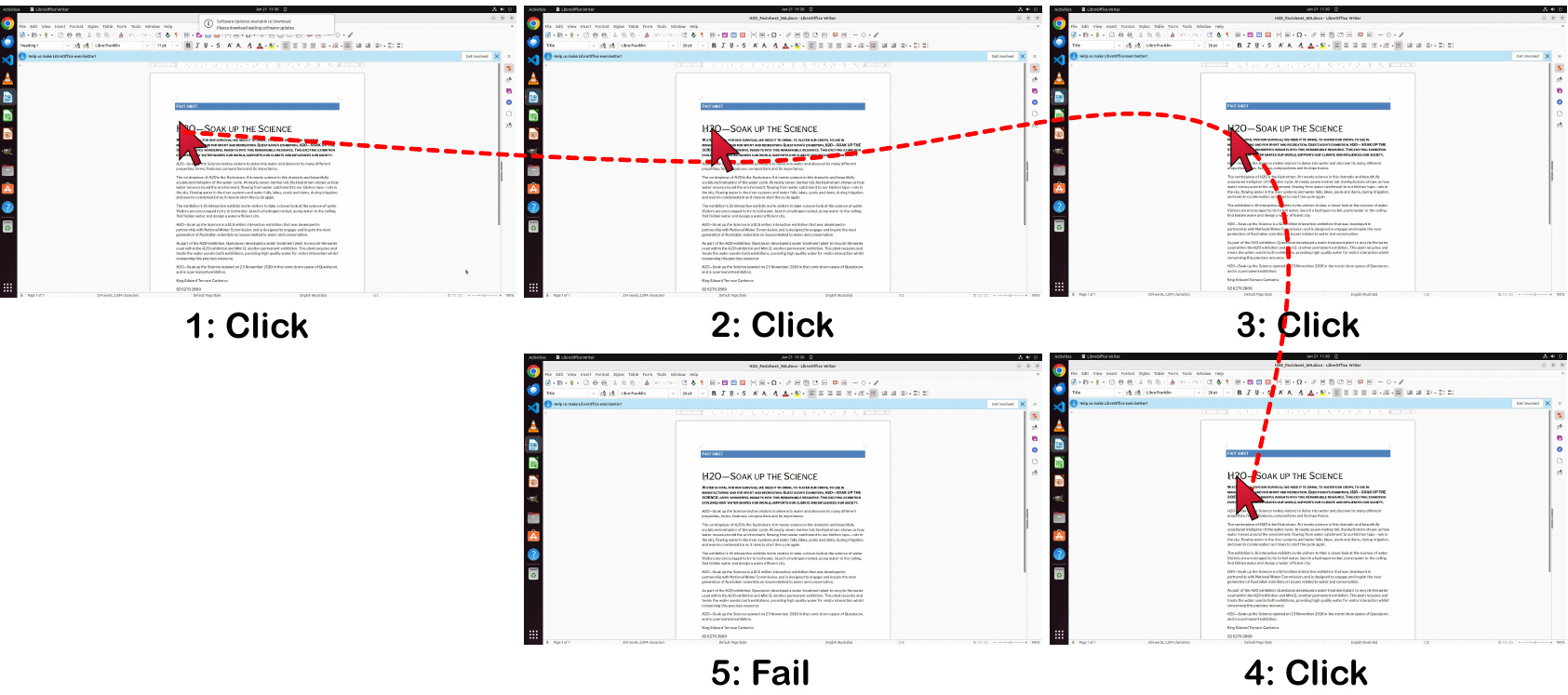}
            \par\vspace{0.25em}
            {\footnotesize \textbf{Top:} Failed trajectory with the conventional reflector.}
        \end{minipage}

        \vspace{0.6em}

        \begin{minipage}{\textwidth}
            \centering
            \includegraphics[width=\linewidth]{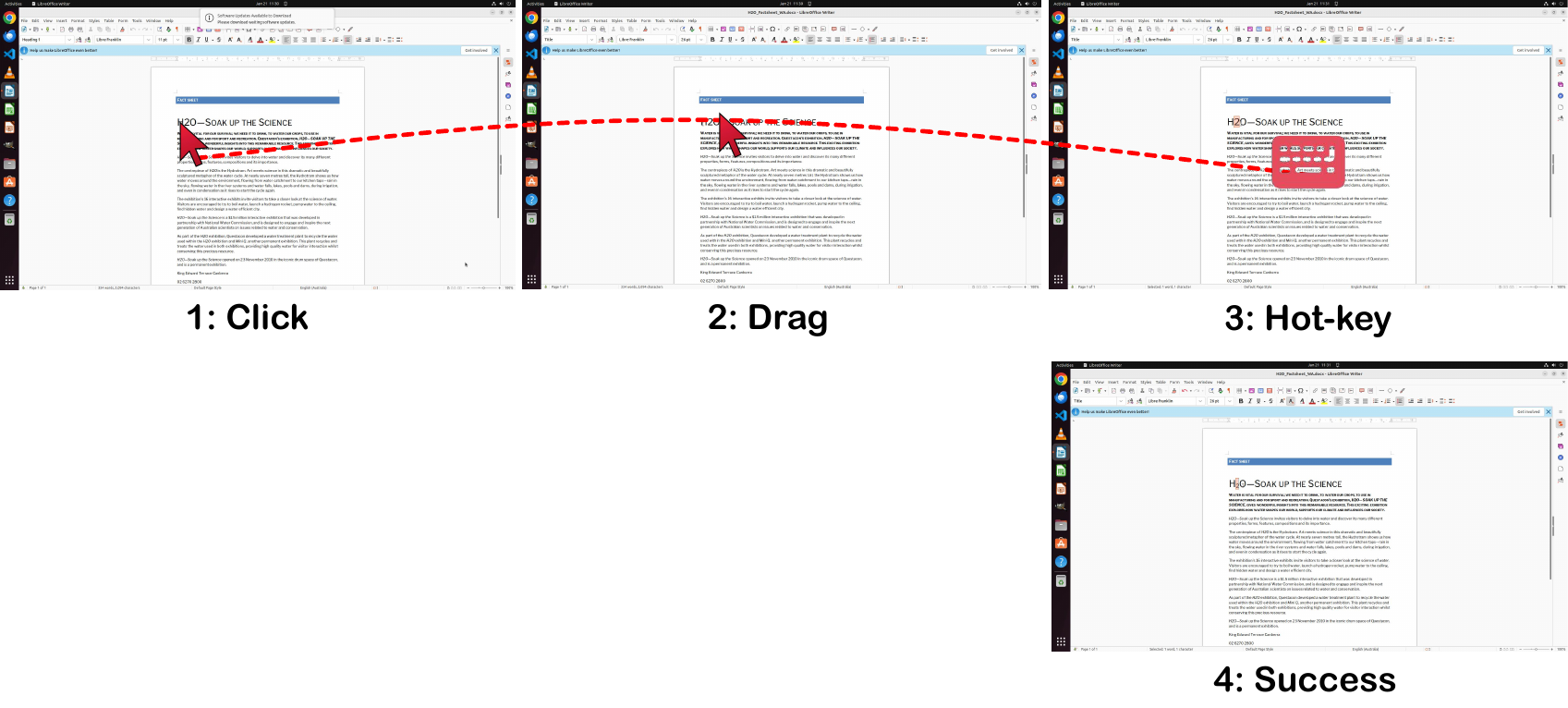}
            \par\vspace{0.25em}
            {\footnotesize \textbf{Bottom:} Successful trajectory with EFR.}
        \end{minipage}
        \captionof{figure}{Case 1: EFR succeeds by recognizing subtle cursor and formatting evidence, while the conventional reflector repeatedly misses the small change and keeps retrying the same click.}
        \label{fig:case-study-writer}
    \end{center}

    \textbf{Failed trajectory with the conventional reflector.}
    \begin{enumerate}[leftmargin=*, labelsep=0.5em]
        \item \textbf{Action:} Click the character ``2''.\\
        \textbf{Reflector:} No visible page change is reported, so the click is judged as failed. However, a small cursor occurs next to character ``2''.
        \item \textbf{Action:} Click the character ``2'' again.\\
        \textbf{Reflector:} The reflector still reports no visible page change and again judges the action as failed. However, a small cursor occurs next to character ``2''.
        \item \textbf{Action:} Click the character ``2'' for a third time.\\
        \textbf{Reflector:} The reflector still reports no visible page change and again judges the action as failed. However, a small cursor occurs next to character ``2''.
        \item \textbf{Action:} Click the character ``2'' for a fourth time.\\
        \textbf{Reflector:} The reflector again reports no visible change, the trajectory remains stuck in repeated retries, and the task ultimately fails.
    \end{enumerate}

    \textbf{Successful trajectory with EFR.}
    \begin{enumerate}[leftmargin=*, labelsep=0.5em]
        \item \textbf{Action:} Click the character ``2''.\\
        \textbf{Reflector:} A text cursor appears next to ``2'', indicating that the click has landed on the correct position.
        \item \textbf{Action:} Drag over the character ``2''.\\
        \textbf{Reflector:} The background color of ``2'' changes, showing that the character has been successfully selected.
        \item \textbf{Action:} Press \texttt{Ctrl+Shift+B}.\\
        \textbf{Reflector:} The ``2'' is rendered as a subscript, so the reflector marks the task as successful.
    \end{enumerate}

    \subsubsection{Case 2: Add Wednesday 12:00 course slot in Calc schedule}

    This case corresponds to \path{multi_apps/3a93cae4-ad3e-403e-8c12-65303b271818}. The task is to enter the course content into the Wednesday 12:00 slot in a Calc spreadsheet. Here the critical difficulty is not a missing change, but a misleading partial success: the target cell is updated, yet unrelated cells are also modified. This example shows how EFR improves the reflector in the large-change regime: instead of checking only whether the intended region contains the expected changes, it first organizes the full set of differences and can therefore reject actions with harmful side effects.

    \begin{center}
        \begin{minipage}{\textwidth}
            \centering
            \includegraphics[width=\linewidth]{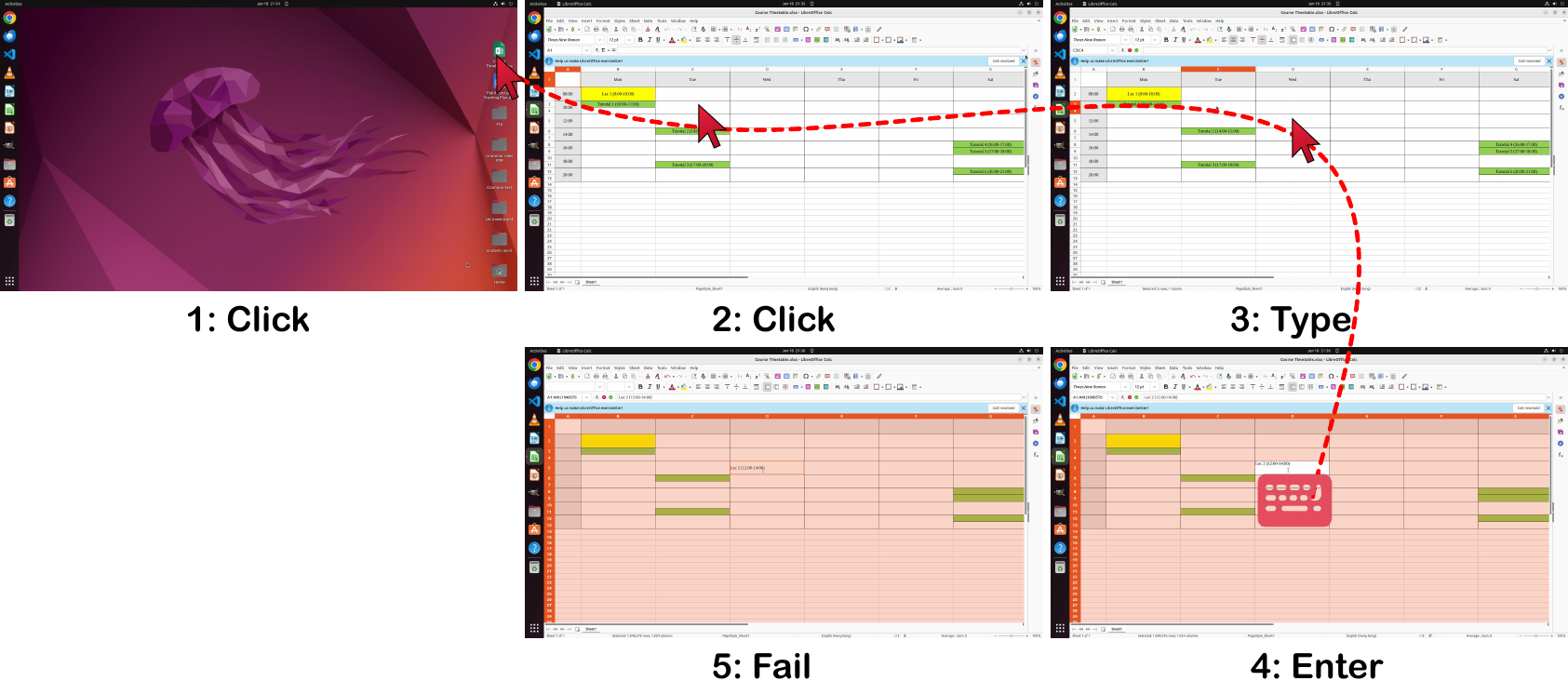}
            \par\vspace{0.25em}
            {\footnotesize \textbf{Top:} Failed trajectory with the conventional reflector.}
        \end{minipage}

        \vspace{0.6em}

        \begin{minipage}{\textwidth}
            \centering
            \includegraphics[width=\linewidth]{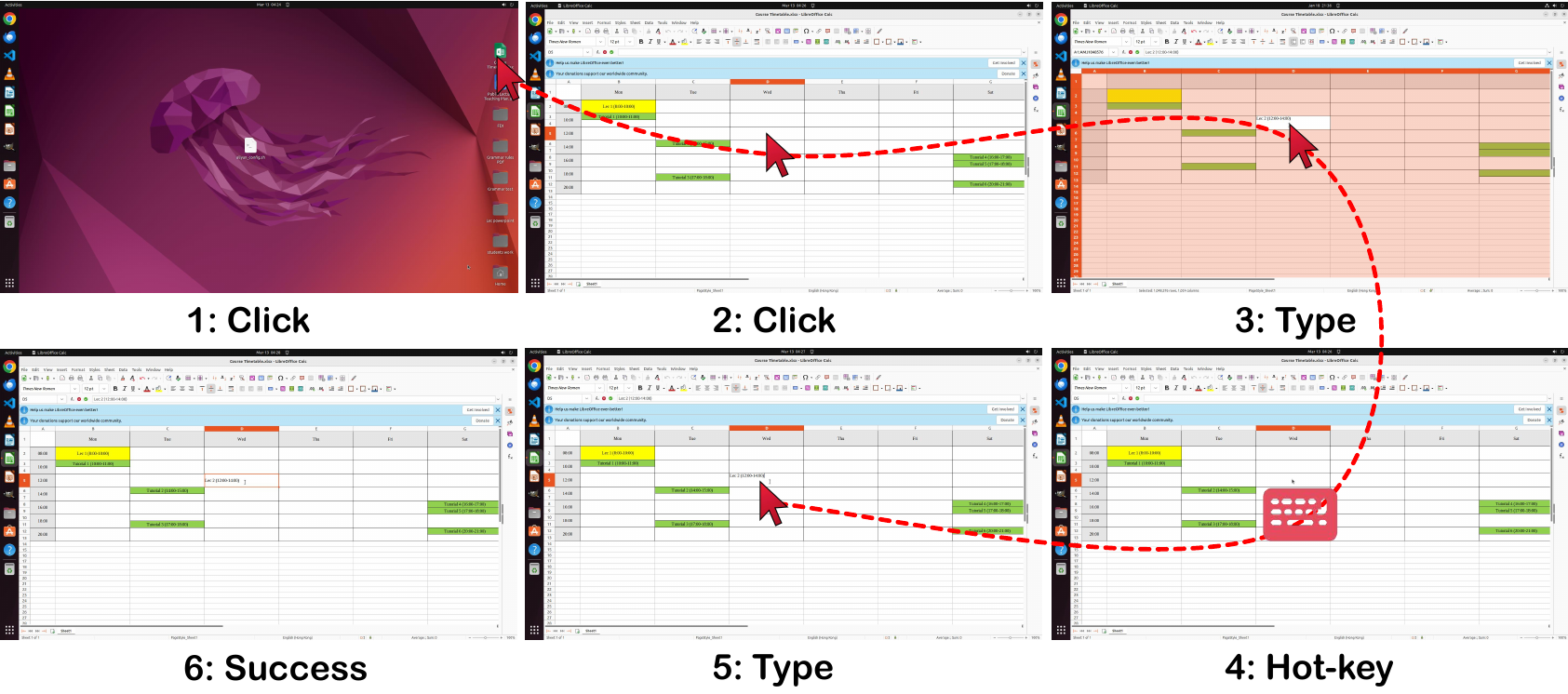}
            \par\vspace{0.25em}
            {\footnotesize \textbf{Bottom:} Successful trajectory with EFR.}
        \end{minipage}
        \captionof{figure}{Case 2: EFR rejects a target-cell update that also damages other cells, triggers recovery, and succeeds on retry; the conventional reflector accepts the same corrupted intermediate state as success.}
        \label{fig:case-study-calc}
    \end{center}

    \textbf{Failed trajectory with the conventional reflector.}
    \begin{enumerate}[leftmargin=*, labelsep=0.5em]
        \item \textbf{Action:} Click the target \texttt{xlsx} file.\\
        \textbf{Reflector:} The page switches to the spreadsheet, confirming that the file is open.
        \item \textbf{Action:} Click the Wednesday 12:00 cell.\\
        \textbf{Reflector:} The changed border indicates that the target cell has been selected.
        \item \textbf{Action:} Select the cell content, delete the original text, and type the new course content.\\
        \textbf{Reflector:} The reflector focuses only on the target cell, observes that the expected content appears there, and therefore judges the action as successful, while overlooking unintended changes in other cells. However, the contents of other cells are deleted as well.
        \item \textbf{Action:} Press \texttt{Enter} to confirm the edit.\\
        \textbf{Reflector:} The target cell exits edit mode, so the trajectory appears successful locally, even though the spreadsheet has already been corrupted globally.
    \end{enumerate}

    \textbf{Successful trajectory with EFR.}
    \begin{enumerate}[leftmargin=*, labelsep=0.5em]
        \item \textbf{Action:} Click the target \texttt{xlsx} file.\\
        \textbf{Reflector:} The page switches to the spreadsheet, confirming that the file has been opened successfully.
        \item \textbf{Action:} Click the Wednesday 12:00 cell.\\
        \textbf{Reflector:} The cell border changes, indicating that the target cell has been selected.
        \item \textbf{Action:} Select the cell content, delete the original text, and type the new course content.\\
        \textbf{Reflector:} Although the target cell is filled with the desired content, the reflector also observes large-scale page changes and missing text in unrelated regions, so this step is judged as failed because the action has harmful side effects.
        \item \textbf{Action:} Undo the previous edit.\\
        \textbf{Reflector:} The spreadsheet returns to the earlier state, confirming that the incorrect edit has been rolled back successfully.
        \item \textbf{Action:} Re-enter the course content in the target cell.\\
        \textbf{Reflector:} This time only the intended cell changes, so the reflector judges the action as successful and the task is completed.
    \end{enumerate}

    \subsection{Failure to Failure}

    In this section, we examine a trajectory pair where both the conventional reflector and EFR fail from OSWorld-Verified. This type of case is important because it clarifies the scope of reflector improvement: EFR can make reflection more evidence-aware, but the reflector is still only one module in a larger GUI-agent system. The essential reason for failure is the errors in the planner and operator. Even if the reflector produces a correct judgment, the subsequent execution trajectory does not necessarily change as a result.

    \subsubsection{Case 3: Move the table on Page 3 to the bottom of the slide}

    This case corresponds to \path{libreoffice_impress/ac1b39ff-ee4d-4483-abce-c117e98942f0}. The task is to move the table on Page 3 to the bottom of the slide. Both trajectories eventually fail, but they reveal different limitations. The conventional reflector mostly accepts each apparent movement as successful, while EFR detects one failed drag action. However, after detecting the failure, the overall agent does not roll back effectively, and later actions still suffer from grounding errors.

    \begin{center}
        \begin{minipage}{\textwidth}
            \centering
            \includegraphics[width=\linewidth]{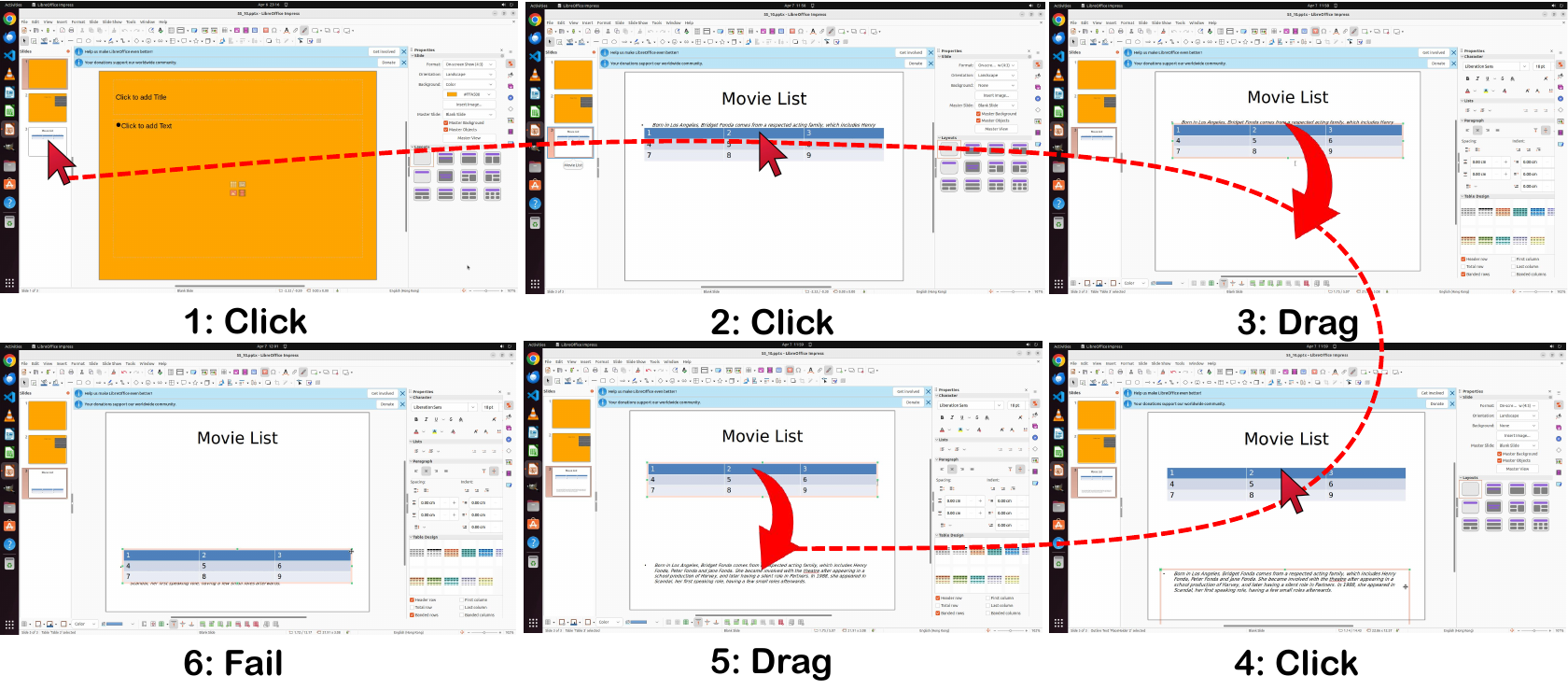}
            \par\vspace{0.25em}
            {\footnotesize \textbf{Top:} Failed trajectory with the conventional reflector.}
        \end{minipage}

        \vspace{0.6em}

        \begin{minipage}{\textwidth}
            \centering
            \includegraphics[width=\linewidth]{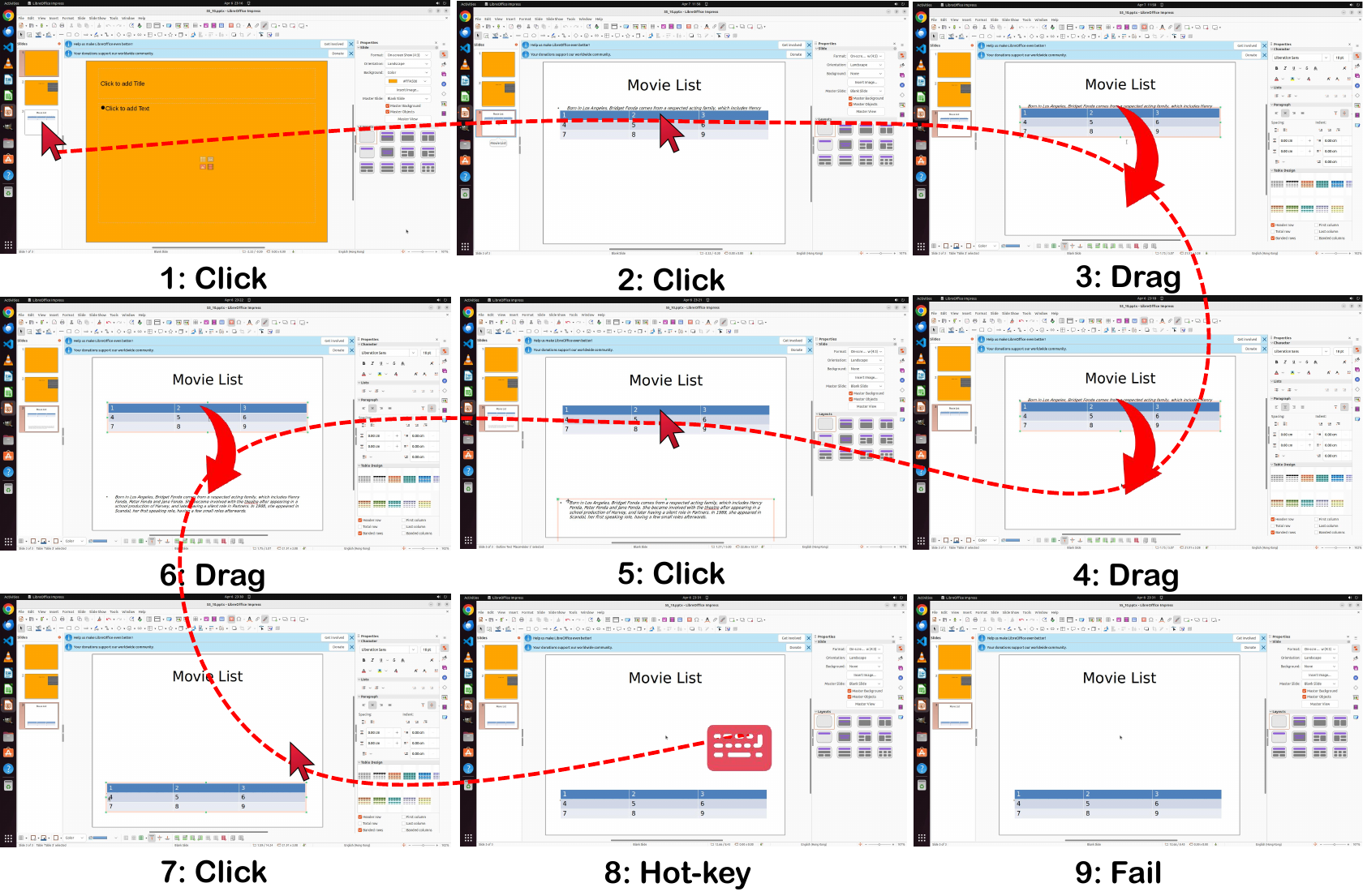}
            \par\vspace{0.25em}
            {\footnotesize \textbf{Bottom:} Failed trajectory with EFR.}
        \end{minipage}
        \captionof{figure}{Case 3: Both reflectors fail to complete the slide-editing task. EFR identifies an unsuccessful drag action, but the agent still does not roll back or re-plan correctly, and grounding errors continue to affect later steps.}
        \label{fig:case-study-impress}
    \end{center}

    \textbf{Failed trajectory with the conventional reflector.}
    \begin{enumerate}[leftmargin=*, labelsep=0.5em]
        \item \textbf{Action:} Click Page 3 in the left slide thumbnail panel.\\
        \textbf{Reflector:} The main canvas switches to Page 3, indicating that the target slide has been opened successfully.
        \item \textbf{Action:} Click the table to select it.\\
        \textbf{Reflector:} Green control points appear around the table, showing that the object has been selected.
        \item \textbf{Action:} Drag the selected object downward.\\
        \textbf{Reflector:} The object appears to move down, so the reflector judges the drag as successful. However, the table is still above the text, and the actual manipulation is likely applied to the wrong object or an incorrect selection region.
        \item \textbf{Action:} Click the table again to adjust its position.\\
        \textbf{Reflector:} Green control points appear again, showing that the object has been selected.
        \item \textbf{Action:} Drag the object below the text.\\
        \textbf{Reflector:} The object moves further toward the bottom area, action successful.
    \end{enumerate}

    \textbf{Failed trajectory with EFR.}
    \begin{enumerate}[leftmargin=*, labelsep=0.5em]
        \item \textbf{Action:} Click Page 3 in the left slide thumbnail panel.\\
        \textbf{Reflector:} The main canvas switches to Page 3, indicating that the target slide has been opened successfully.
        \item \textbf{Action:} Click the table to select it.\\
        \textbf{Reflector:} Green control points appear around the table, showing that the object has been selected.
        \item \textbf{Action:} Drag the selected object downward.\\
        \textbf{Reflector:} The page does not visibly change, action failed.
        \item \textbf{Action:} Try again to drag the selected object downward.\\
        \textbf{Reflector:} The text appears to move down instead of the table, so the reflector judges the drag as failed. However, the agent does not roll back. Instead, it continues with the next steps, which still suffer from grounding errors.
        \item \textbf{Action:} Click the table again to adjust the position.\\
        \textbf{Reflector:} The table is selected and green control points appear, action successful.
        \item \textbf{Action:} Continue dragging the table below the text.\\
        \textbf{Reflector:} The selected object moves further toward the bottom region, action successful.
        \item \textbf{Action:} Click an empty area of the slide.\\
        \textbf{Reflector:} The table selection is canceled successfully.
        \item \textbf{Action:} Press \texttt{Ctrl+S} to save the file.\\
        \textbf{Reflector:} The file is saved successfully, but the saved state still contains the unresolved layout error.
    \end{enumerate}

    \subsection{Success to Failure}
    \label{sec:success-to-fail}

    In this section, we examine a negative transition trajectory introduced by EFR from OSWorld-Verified. This example shows that the current reflector can still make an incorrect final judgment even when it observes the key visual evidence, revealing that success judgment remains difficult even when visual evidence is sufficient.

    \subsubsection{Case 4: Place the photo on the desktop and rename it to \texttt{export.jpg}}    
    This case corresponds to \path{gimp/77b8ab4d-994f-43ac-8930-8ca087d7c4b4}. The task is to export the opened image to the desktop and rename it as \texttt{export.jpg}. The conventional reflector completes the export correctly, while EFR fails because it accepts \texttt{export.jpg.jpg} as a successful filename. Notably, the original trajectory succeeds because its operator takes the correct action, rather than because the conventional reflector is intrinsically better than EFR. This example highlights a remaining limitation of the reflector: EFR can improve evidence extraction, but the final success judgment still depends on correctly interpreting the semantics of the observed evidence.

    \begin{center}
        \begin{minipage}{\textwidth}
            \centering
            \includegraphics[width=\linewidth]{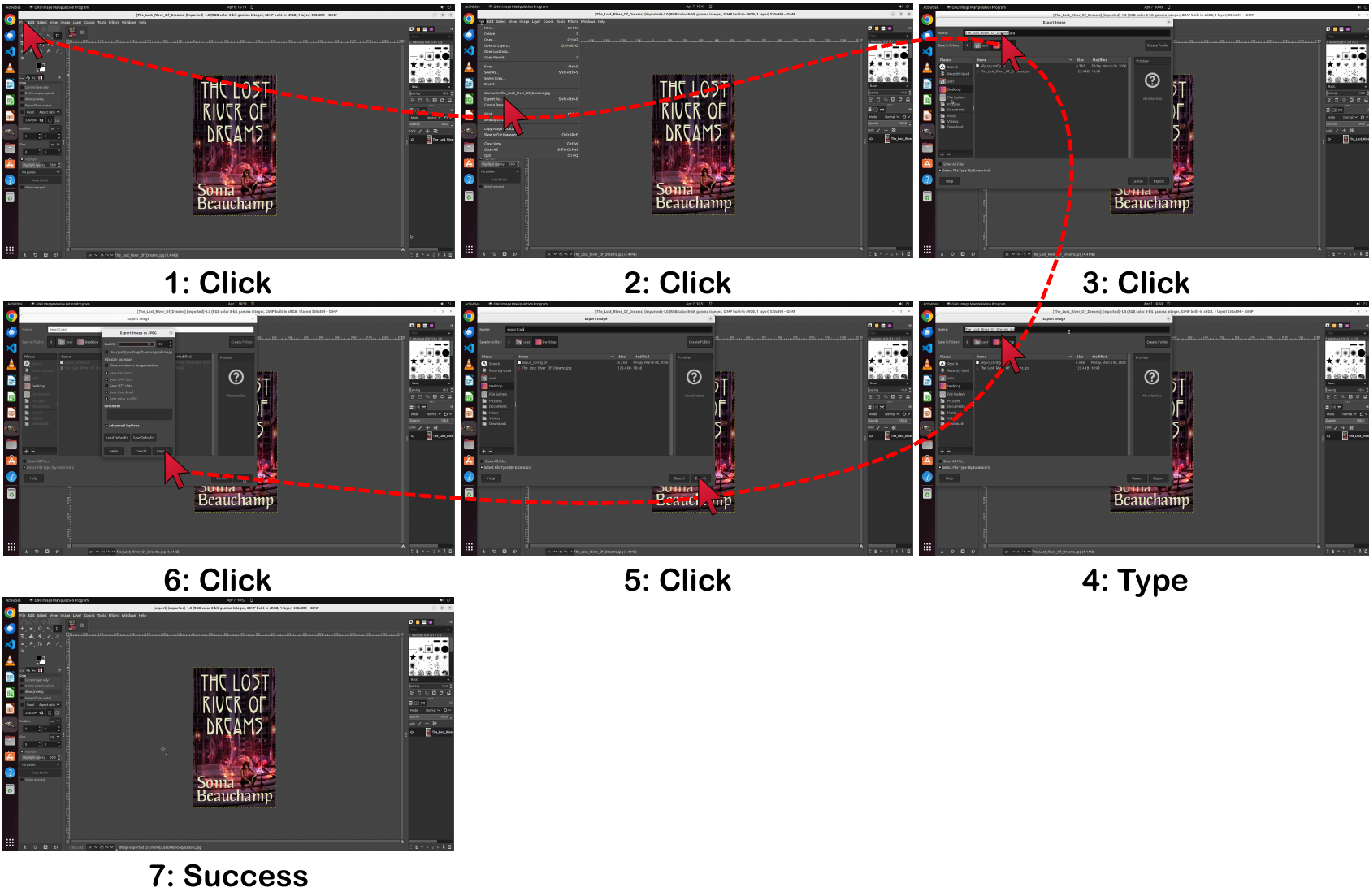}
            \par\vspace{0.25em}
            {\footnotesize \textbf{Top:} Successful trajectory with the conventional reflector.}
        \end{minipage}

        \vspace{0.6em}

        \begin{minipage}{\textwidth}
            \centering
            \includegraphics[width=\linewidth]{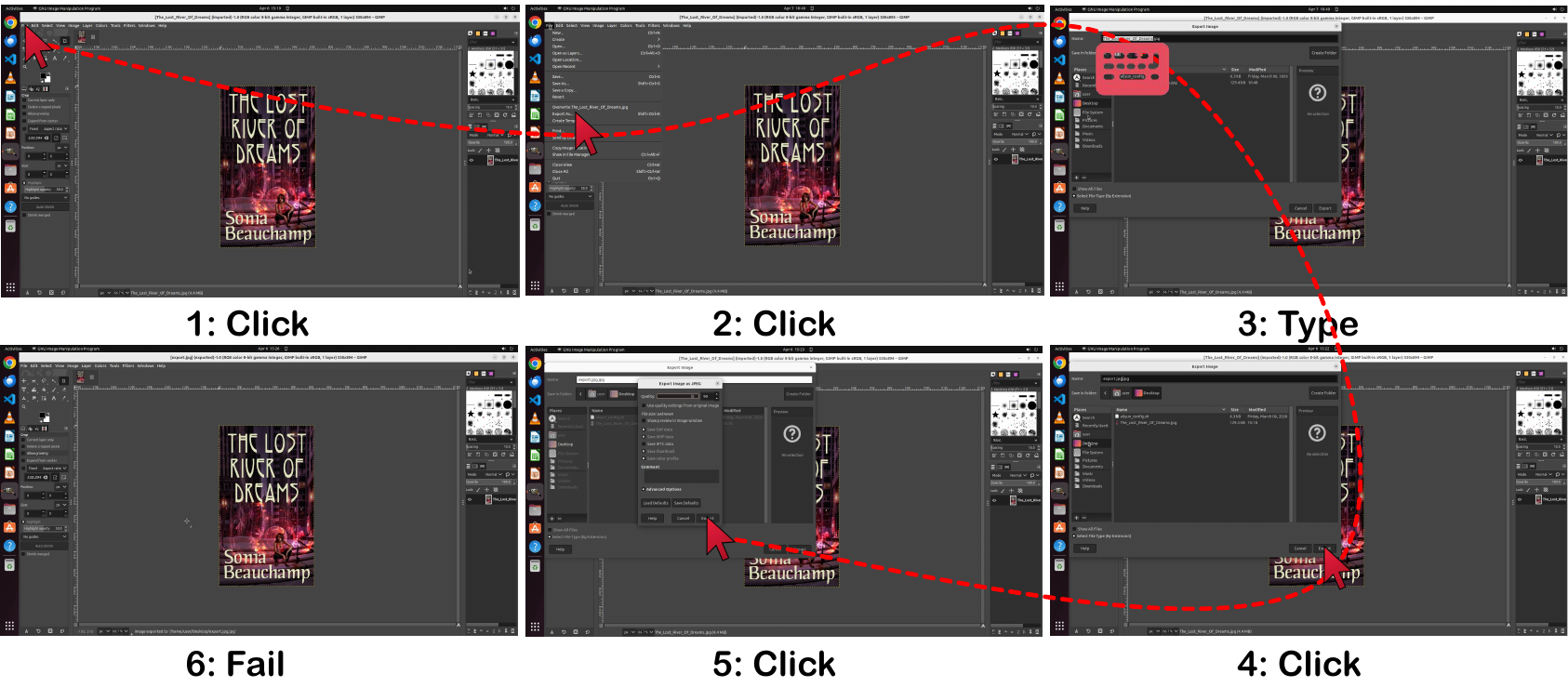}
            \par\vspace{0.25em}
            {\footnotesize \textbf{Bottom:} Failed trajectory with EFR.}
        \end{minipage}
        \captionof{figure}{Case 4: The conventional reflector exports the image as \texttt{export.jpg}, while EFR observes the duplicated suffix \texttt{export.jpg.jpg} but incorrectly judges it as harmless.}
        \label{fig:case-study-gimp}
    \end{center}

    \textbf{Successful trajectory with the conventional reflector.}
    \begin{enumerate}[leftmargin=*, labelsep=0.5em]
        \item \textbf{Action:} Click \texttt{File}.\\
        \textbf{Reflector:} A drop-down menu appears, indicating that the click succeeds.
        \item \textbf{Action:} Click \texttt{Export As}.\\
        \textbf{Reflector:} The export configuration dialog appears, confirming that the export operation has been opened successfully.
        \item \textbf{Action:} Triple-click the filename input box.\\
        \textbf{Reflector:} The current filename is selected, so the input field is ready to be replaced.
        \item \textbf{Action:} Type \texttt{export.jpg}.\\
        \textbf{Reflector:} The filename changes to \texttt{export.jpg}, indicating that the rename step succeeds.
        \item \textbf{Action:} Click \texttt{Export}.\\
        \textbf{Reflector:} A new export configuration dialog appears, confirming that the first export click succeeds.
        \item \textbf{Action:} Click \texttt{Export} again.\\
        \textbf{Reflector:} The export configuration dialog disappears, indicating that the export has been confirmed and the task is completed successfully.
    \end{enumerate}

    \textbf{Failed trajectory with EFR.}
    \begin{enumerate}[leftmargin=*, labelsep=0.5em]
        \item \textbf{Action:} Click \texttt{File}.\\
        \textbf{Reflector:} A drop-down menu appears, indicating that the click succeeds.
        \item \textbf{Action:} Click \texttt{Export As}.\\
        \textbf{Reflector:} The export configuration dialog appears, confirming that the export operation has been opened successfully.
        \item \textbf{Action:} Type \texttt{export.jpg} in the filename input box.\\
        \textbf{Reflector:} The filename text is selected, so the reflector judges that the filename field has been activated successfully.
        \item \textbf{Action:} Type \texttt{export.jpg}.\\
        \textbf{Reflector:} The filename becomes \texttt{export.jpg.jpg}. EFR correctly observes the duplicated suffix, but incorrectly judges that the extra suffix is caused by system behavior and does not affect the task.
        \item \textbf{Action:} Click \texttt{Export}.\\
        \textbf{Reflector:} A new export configuration dialog appears, confirming that the first export click succeeds.
        \item \textbf{Action:} Click \texttt{Export} again.\\
        \textbf{Reflector:} The export configuration dialog disappears, so the reflector judges the export as complete, but the final filename is still incorrect.
    \end{enumerate}

    \subsection{EFR Failure Case Analysis}
    \label{sec:efr-failure-analysis}

    The two EFR failure cases above reflect two different sources of remaining errors. In Case~3, the main issue is fine-grained grounding. Moving an object in a slide requires accurately selecting the intended table and dragging it to a precise location. EFR detects that one drag action does not produce the intended effect, and later observes that the text rather than the table appears to move. However, this reflection signal is not effectively converted into a recovery strategy by the subsequent planner and operator. The agent continues to issue similar manipulation actions, so the trajectory still accumulates unresolved layout errors. This case shows that even when the reflector correctly identifies a local execution failure, the full agent may still fail if later modules cannot revise the plan or repair the state.

    In Case~4, the failure comes from both an input error and a semantic misunderstanding of the observed change. The agent first fails to properly replace the existing filename, causing the typed string to produce \texttt{export.jpg.jpg}. EFR observes this duplicated suffix, but incorrectly interprets it as a harmless system behavior rather than a violation of the target filename requirement. As a result, the agent accepts the intermediate state and proceeds to export the file with an incorrect name. This case shows that explicit visual evidence alone is not always sufficient: the final judgment still requires correctly mapping observed changes to task semantics, especially when small textual differences determine task success.

    Overall, these cases indicate that the scattered negative results are not caused by a systematic weakness of EFR on a particular task category. Instead, they arise from limitations of the broader agent pipeline: precise grounding and manipulation remain difficult, recovery from harmful actions is still weak, and the verifier can still misinterpret the semantics of correctly observed evidence.


\end{document}